\documentclass{article}

\usepackage{arxiv}
\usepackage{verbatim}
\usepackage[utf8]{inputenc} 
\usepackage[T1]{fontenc}    
\usepackage{hyperref}       
\usepackage{url}            
\usepackage{booktabs}       
\usepackage{amsfonts}       
\usepackage{nicefrac}       
\usepackage{microtype}      
\usepackage{lipsum}		
\usepackage{graphicx}
\usepackage{natbib}
\usepackage{doi}
\usepackage{subfigure}
\usepackage{amsmath}

\title{Abutting Grating Illusion: Cognitive Challenge to Neural Network Models}

\author{{Jinyu Fan\,$^{1}$, Yi Zeng\,$^{1,2,3,4}$\thanks{Corresponding author: jinyu.fan@ia.ac.cn}} \\
	$^{1}$Research Center for Brain-Inspired Intelligence, Institute of Automation, Chinese Academy of Sciences, Beijing, China, \\
$^{2}$Center for Excellence in Brain Science and Intelligence Technology,Chinese Academy of Sciences, Shanghai, China \\
$^{3}$School of Artificial Intelligence, University of Chinese Academy of Sciences, Beijing, China,\\
$^{4}$National Laboratory of Pattern Recognition, Institute of Automation,Chinese Academy of Sciences, Beijing, China
}

\hypersetup{
pdftitle={A template for the arxiv style},
pdfsubject={q-bio.NC, q-bio.QM},
pdfauthor={David S.~Hippocampus, Elias D.~Striatum},
pdfkeywords={First keyword, Second keyword, More},
}

\begin{document}
\maketitle

\begin{abstract}
Even the state-of-the-art deep learning models lack fundamental abilities compared to humans. Multiple comparison paradigms have been proposed to explore the distinctions between humans and deep learning. While most comparisons rely on corruptions inspired by mathematical transformations, very few have bases on human cognitive phenomena. In this study, we propose a novel corruption method based on the abutting grating illusion, which is a visual phenomenon widely discovered in both human and a wide range of animal species. The corruption method destroys the gradient-defined boundaries and generates the perception of illusory contours using line gratings abutting each other. We applied the method on MNIST, high resolution MNIST, and silhouette object images. Various deep learning models are tested on the corruption, including models trained from scratch and 109 models pretrained with ImageNet or various data augmentation techniques. Our results show that abutting grating corruption is challenging even for state-of-the-art deep learning models because most models are randomly guessing. We also discovered that the DeepAugment technique can greatly improve robustness against abutting grating illusion. Visualisation of early layers indicates that better performing models exhibit stronger end-stopping property, which is consistent with neuroscience discoveries. To validate the corruption method, 24 human subjects are involved to classify samples of corrupted datasets. 
\end{abstract}


\section{Introduction}

Deep learning has achieved tremendous success during the past decade, even surpassing human performance in numerous vision tasks\citep{russakovsky2015imagenet}\citep{dodge2017study}. However, it is certainly not the panicillin to all vision tasks that humans can perform. While ANN models can achieve extremely high results on test set drawn from the same distribution of training set, they can easily fail facing with OOD(out-of-distribution) data\citep{dodge2017study}. It has been 
discovered that neural network performance decreases under different image corruptions, such as noise, blur, brightness change, fog, etc\citep{dodge2016understanding}\citep{hendrycks2019benchmarking}. On the other hand, humans are extremely robust to different sorts of distortions applied to images\citep{dodge2017study}. An even more extreme case is adversarial attacks, where human-imperceptible perturbations could 
cause catastrophic failures to well-trained neural network models\citep{szegedy2013intriguing}. Multiple attack and defense mechanisms\citep{szegedy2013intriguing}\citep{carlini2017towards}\citep{madry2017towards}\citep{moosavi2016deepfool}\citep{papernot2016distillation} have been proposed in recent years, but the problem still remains unsolved. Moreover, it has been found that the errors made by humans and nerual network models have little correlation with each other\citep{dodge2017study}, indicating that current machine visual systems might still have fundamental deficits compared to human visual systems.

Comparing human and machine visual systems under different corruptions is certainly not a brand new idea. Multiple experiments have been conducted between deep learning models and humans to explore the weakness of neural network models\citep{borji2014human}\citep{fleuret2011comparing}\citep{stabinger201625}\citep{parikh2011recognizing}\citep{dodge2017study}. It has been proved that humans are more robust against different corruptions than neural networks. Most corruptions used are noise naturally occurring in daily life\citep{hendrycks2021many} or simply mathematical transformations defined in image processing. A crucial problem is that they seldom mention or discuss the relationship between the proposed distortions and human cognitive functions. Thus, even if engineering methods such as data augmentation\citep{ford2019adversarial}\citep{lopes2019improving}\citep{hendrycks2019augmix}\citep{rusak2020increasing}, adversarial training\citep{goodfellow2014explaining}\citep{madry2017towards}\citep{laidlaw2020perceptual} and pretraining\citep{hendrycks2019using} are proposed to improve the model robustness, they might not be helpful to understand why humans are robust to different corruptions and how to gain inspirations from biological mechanisms to improve deep learning models.

On the other hand, there are also corruptions that draw inspirations from psychology studies. Some researchers explored the texture vs shape bias problems in deep learning models. Shape bias has been found in human object recognition that shape cues play the most crucial role in object recognition of both young children and adults compared to other features such as colors or sizes\citep{landau1988importance}\citep{gershkoff2004shape}. Whether neural networks are shape-biased or texture-biased still remains an open question. Recent studies suggest that deep learning models are texture-biased\citep{hosseini2018assessing}\citep{gatys2017texture}\citep{brendel2019approximating}\citep{geirhos2018imagenet}\citep{hermann2020origins}, although there are also contradictory evidences supporting the other side of the argument\citep{kubilius2016deep}\citep{ritter2017cognitive}. To study the shape bias problem, different approaches have been proposed to decouple the shape and texture information\citep{kubilius2016deep}\citep{borji2014human}\citep{geirhos2018imagenet}\citep{geirhos2018imagenet}\citep{hermann2020origins}\citep{hosseini2018assessing}\citep{gatys2017texture}\citep{brendel2019approximating}\citep{borji2014human}\citep{geirhos2018imagenet}. In these studies, when the texture information is corrupted, 
both global shape and local edge information is usually maintained as clear boundaries of luminance contrast. On the other hand, in order to corrupt the shape information, the common strategy is to scramble the spatial positions of local areas, so that the global shape information is destroyed while local information, including local edges and local textures, are preserved. 

\begin{figure}[htbp]
\centering
\includegraphics[scale=0.6]{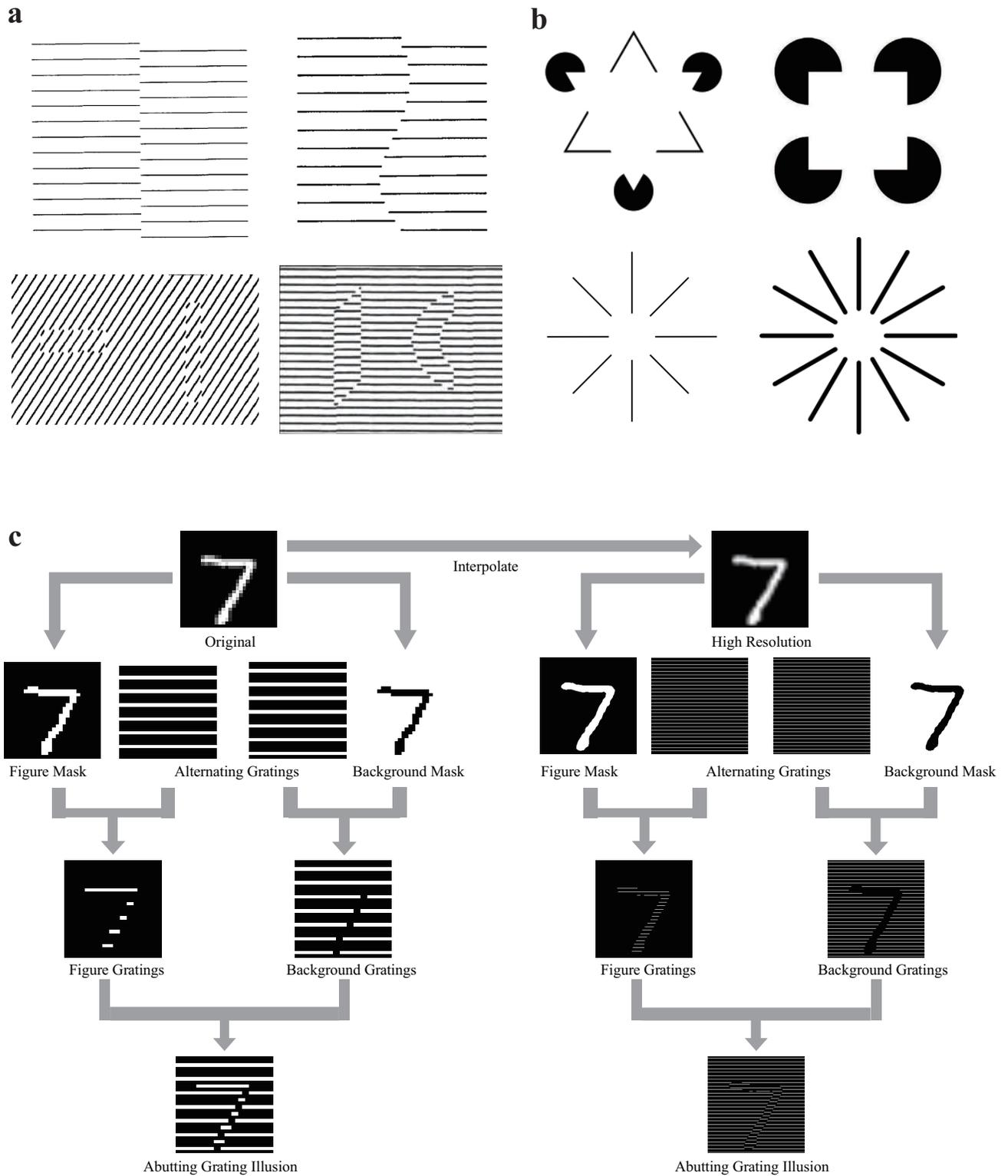}
\caption{\textbf{a}, Examples of abutting grating illusions\citep{von1989mechanisms}\citep{gurnsey1992parallel}\citep{davis1998kanizsa}. \textbf{b}, Examples of Kanizsa triangle, Kanizsa square and Erhenstein illusion\citep{kanizsa1976subjective}\citep{ehrenstein1941abwandlungen}. \textbf{c}, Procedure of converting MNIST sample to abutting grating MNIST sample and high-resolution abutting grating MNIST.}
\label{Fig:abutting}
\end{figure}

In fact, human visual systems are so powerful that we are capable of recognizing objects even with neither physical boundaries of luminance contrast, nor local texture and local edge information at all. Abutting grating illusion\citep{kanizsa1974contours} is a classic visual illusion where displaced line gratings induce the perception of illusory edges and shapes in the absence of luminance contrast. The upper left image of Fig.\ref{Fig:abutting}a demonstrates the standard pattern of abutting grating illusion, where an illusory vertical line could be perceived although it does not possess a physical boundary. More variations of abutting grating illusion have been created in following studies as demonstrated in Fig.\ref{Fig:abutting}a\citep{von1989mechanisms}\citep{gurnsey1992parallel}\citep{davis1998kanizsa}. The abutting grating illusion is one type of illusory contours, also known as subjective contours, which was first discovered by Friedrich Schumann around the beginning of the 20th century\citep{schumann1918beitrage}. Illusory contours are visual illusions that evoke the perception of a clear boundary without color contrast or luminance gradients across that boundary. A very distinguishable feature of illusory contours is that the local edge information is completely destroyed. Applying classic edge detectors such as canny edge detector\citep{canny1986computational} cannot produce the illusory boundaries. On the contrary, humans can easily perceive the global existence of an apparent vertical line at the middle. The abutting grating illusion is one of the illusory contour examples that have been widely used in physiological studies to explore biological visual processing of illusory contours\citep{pan2012equivalent}\citep{de1996cue}\citep{montaser2007orientation}\citep{ramsden2001real}. Except for the abutting grating illusion, there are also other types of illusory contours proposed in the last century, such as Kanizsa triangle\citep{kanizsa1976subjective} and Erhenstein illusion\citep{ehrenstein1941abwandlungen} as in Fig.\ref{Fig:abutting}b.

Compared to the common corruptions or shape bias problem, whether deep learning models can perceive illusory contours is a relatively under explored area, with very few researches studying this topic. \citep{baker2018deep}\citep{kellman2017classification} found that CNN models failed to reveal behavioral receptive field activity along illusory contours using the "fat/thin" task\citep{ringach1996spatial} which was originally designed to study the spatial and temporal properties of illusory contours in humans. On the other hand, both \citep{lotter2018neural} and \citep{pang2021predictive} proposed that applying predictive coding\citep{rao1999predictive} to neural network architectures could lead to the ability of perceiving illusory contours. One common strategy of these studies is that they pre-train the deep learning models with natural images such as ImageNet or Cifar100. The model is later fine-tuned on toy datasets of squares and their corruptions. Then the analysis of illusory contour perceiving ability generally relies on Kanizsa squares, a variation of Kanizsa triangle. One major problem of this strategy is that the deep learning models are fine-tuned and tested on simple synthetic binary classification problem, thus we cannot 
measure the illusory contours perceiving ability on more complex real-world tasks such as recognizing hand-written digits or natural images in a more direct and quantitative manner. Furthermore, currently all studies are based on Kanizsa squares, while the other types of illusory contours, including Erhenstein illusion and abutting grating illusion, both of which are also very commonly used in physiological studies, are not considered in these studies.

In this paper, we propose a novel corruption method named abutting grating corruption based on the abutting grating illusion as a tool to quantitatively study how deep learning models react to illusory contours. The corruption method can be applied on silhouette-like images which have only the outer contours without any texture information. In this study, we apply the abutting grating corruption on MNIST and 160 silhouette images from \citep{geirhos2018imagenet}. Moreover, we interpolate the original MNIST image samples of 28x28 to 224x224 which can preserve more details and generate stronger illusion when using abutting grating corruption. Various deep learning models are tested on the abutting grating images. For MNIST and the high resolution version, we trained models from scratch with the original MNIST train set and then test them on the abutting grating corrupted test set. For the 160 silhouette images, we collected 109 publicly available pretrained models, from pytorch and timm packages as well as github, most of which are pretrained with ImageNet or various data augmentation techniques to increase their robustness against diverse corruptions. The results show that the performance of most deep learning models are very close to random guess, except for the DeepAugment data augmentation technique, which exhibit a significant improvement over all other methods. Furthermore, under certain parameter settings, the abutting grating corruptions could cause difficulty to human recognition. Thus, 24 subjects participated the human experiments to provide a reference of human performance on several parameter settings of the corrupted datasets. Finally, we visualise the activation maps of the early layers of better performing models. We discovered a phenomenon very similar to the end-stopping property\citep{hubel1965receptive}\citep{hubel1968receptive} that is considered involved in early  processing of illusory contours in biological visual systems\citep{peterhans1986neuronal}\citep{finkel1989integration}\citep{lesher1995illusory}\citep{peterhans1989mechanisms}\citep{von1989mechanisms}\citep{heitger1994computational}\citep{francis1996cortical}. We further spotted certain convolution filters that resemble the hypothetical topology of end-stopped cells theoretically predicted and extensively used in modelling illusory contours\citep{peterhans1986neuronal}\citep{lesher1995illusory}.

\section{Background}
\subsection{Illusory contours}
Illusory contours have been widely discovered and studied in not only humans, but also different animal species, including mammals, birds and insects\citep{nieder2002seeing}. The fact that the perception of illusory contours is widely discovered in independently evolved visual systems indicates its fundamental and crucial role in biological visual processing. Perceiving illusory contours ought to be an inevitable ability for any artificial visual system.

Besides the abutting grating illusion, there are also other types of illusory contours that have been extensively studied. The most famous example is the Kanizsa triangle\citep{kanizsa1976subjective} as in Fig.\ref{Fig:abutting}b which is created by the Italian psychologist Gaetano Kanizsa in 1955\citep{kanizsa1955margini}. In the Kanizsa triangle, three PacMan shapes induce the perception of a white salient triangle at the middle which appears to be brighter than the surrounding area, even though that triangle has no edges at all and its brightness is the same as the background. A variation of Kanizsa triangle is Kanizsa square, which is also very commonly used in physiological studies\citep{lee2001dynamics}\citep{bakin2000visual}\citep{cox2013receptive}. Another type of illusory contours is the Ehrenstein illusion, depicted in the second row of Fig.\ref{Fig:abutting}b. The Ehrenstein illusion is an illusory figure triggered by several radial line segments which was first described by Walter Ehrenstein senior\citep{ehrenstein1941abwandlungen}. 

The cortical processing of abutting grating illusion and illusory contours still remains unknown. The neural activities for illusory contours appear at relatively early visual brain areas, as previous electrophysiology studies reported that about one third of V2 neurons respond to illusory contours\citep{von1989mechanisms}\citep{von1984illusory} as well as very small portion of V1 neurons\citep{grosof1993macaque}, with responses in V2 stronger and earlier than those in V1\citep{lee2001dynamics}. Moreover, electrophysiology studies show that V4 might play a vital role of illusory contour perception, with \citep{de1996cue} showing that V4 lesions lead to significant deficits in illusory contour perception, and \citep{pan2012equivalent} 
revealing that real and illusory contours are represented equivalently at both population and single-cells levels.

\subsection{Corruptions to decouple shape and texture information}
In order to study the shape bias problem, different methods can be used to decouple the shape information from the texture information. In order to corrupt the texture information, \citep{kubilius2016deep} used silhouettes to eliminate texture information. Line drawings and sketches are tested in \citep{borji2014human} which has no texture information. \citep{geirhos2018imagenet} chose 160 natural colour images of objects with white background, and converted them to silhouettes as well as edges using canny edge extractor. \citep{geirhos2018imagenet} also proposed Stylized-ImageNet(SIN) by applying style transfer\citep{gatys2016image} to ImageNet images so that the shape information is well-preserved while the texture information is replaced by randomly selected paintings that are irrelevant to the true label. \citep{hermann2020origins} took noise type as texture and ImageNet class as shape in ImageNet-C dataset\citep{hendrycks2018benchmarking}. \citep{hosseini2018assessing} modified the texture of MNIST\citep{lecun1998mnist}\citep{deng2012mnist} and notMNIST\citep{bulatov2011notmnist} datasets by computing the negative images. To corrupt the shape information, \citep{gatys2017texture}\citep{brendel2019approximating} used CNN texture model\citep{gatys2015texture} to produce texturised images by scrambling the global shape of the images. Moreover, images are jumbled by variable block size in \citep{borji2014human} so that the global shape is destroyed. These two papers both destroyed the global shape information while the local shape information is maintained. On the contrary, \citep{geirhos2018imagenet} directly tested their models with full-width patches of textures, which has absolutely no shape information.

\section{Methods}
\subsection{Abutting grating illusion with MNIST(AG-MNIST)}
\begin{figure}[htbp]
\centering
\includegraphics[scale=0.38]{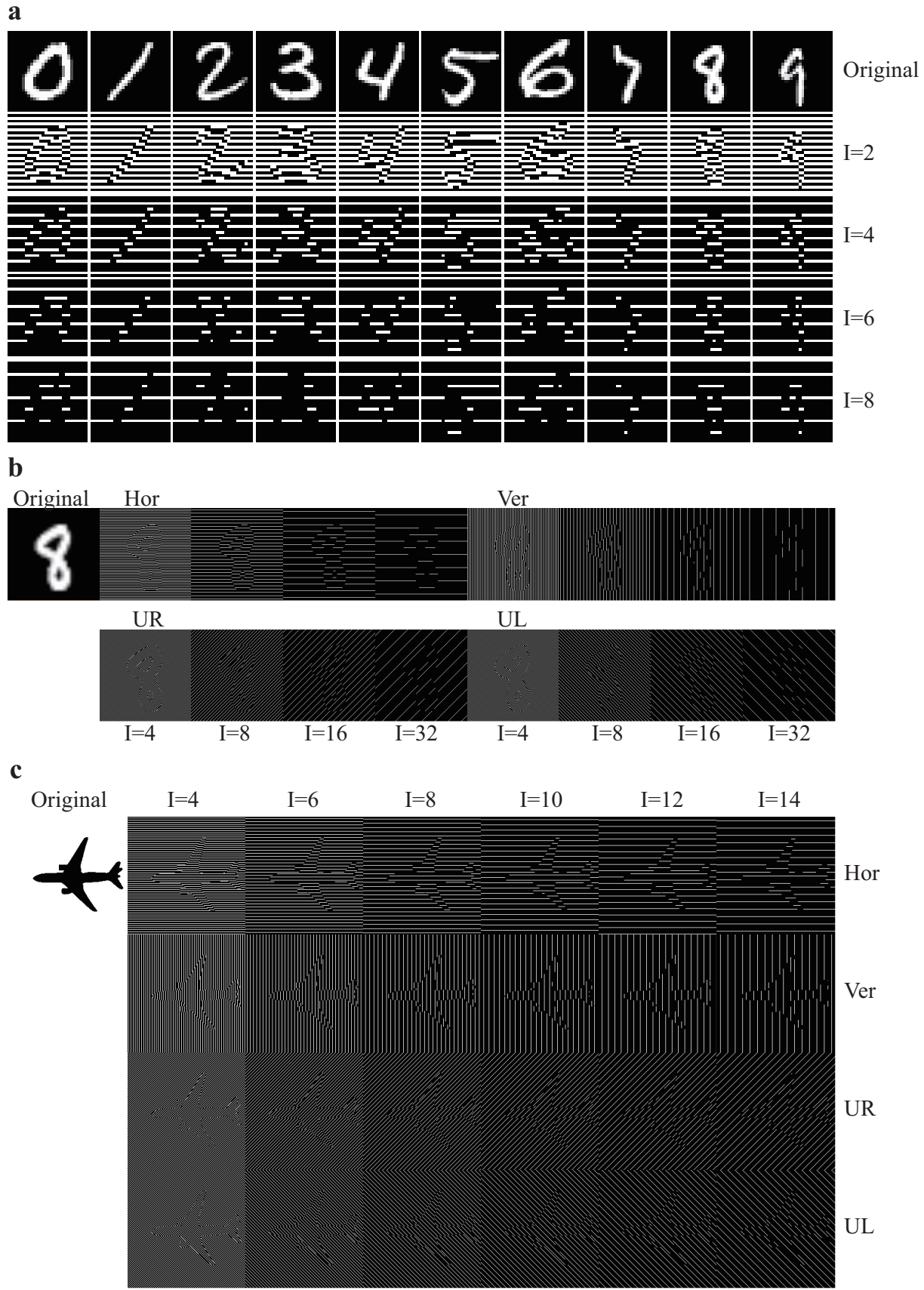}
\caption{\textbf{a}, Demonstration of AG-MNIST, using horizontal gratings of interval 2,4,6 and 8. \textbf{b}, Demonstration of high-resolution AG-MNIST, using horizontal(Hor), vertical(Ver), upper right to lower left(UR) and upper left to lower right(UL) gratings, the interval is set to 4,8,16,32 respectively. \textbf{c}, Demonstration of AG-silhouettes, using horizontal(Hor), vertical(Ver), upper right to lower left(UR) and upper left to lower right(UL) gratings, the interval is set to 4,6,8,10,12,14 respectively.}
\label{Fig:AG-MNIST28}
\end{figure}

A dataset of abutting grating illusion is necessary in order to explore the phenomenon in deep learning models. Images of various labels can easily be collected using a camera or scraping from the Internet for tasks such as recognition, detection or segmentation. However, it is not straightforward how to find samples of abutting grating illusion in natural images or in daily life, because all the original samples of abutting grating illusions in psychology studies are designed manually, let alone large quantities of abutting grating samples with different labels. 
Thus in this study we propose abutting grating corruption, which aims to transform existing vision datasets to their corresponding abutting grating illusion version. One of the critical difficulties of choosing datasets is that abutting grating illusion studies simple contours without texture or minor edge information, while most deep learning vision datasets have very complex features including edges and textures. MNIST dataset is a suitable choice to generate abutting grating illusion due to the fact that numbers are all based on outer contours, and that any potential textures or edges inside the number contour are completely useless information. To create abutting grating illusion with MNIST image samples, we follow the procedure 
shown in Fig.\ref{Fig:abutting}c. The image is first separated into two masks representing the figure and the background respectively by comparing each pixel value to a threshold. Then two images of alternating gratings are generated with the same interval but a phase shift of half a cycle. The grating images are multiplied with the two masks, which produces masked figure gratings and masked background gratings. The two grating images can be joined together to produce the abutting grating illusion version of MNIST image samples. We demonstrate the AG samples induced by horizontal gratings of various intervals in Fig.\ref{Fig:AG-MNIST28}a. The threshold is set to 0.5. We can see that compared to the clear image samples of number figures, the abutting grating illusion version of number is more challenging to perceive even for human beings. With intervals growing larger, more information is destroyed, leading to more difficulties to recognize the figure. The most crucial problem is that the original MNIST image samples are of very low resolution, which allows only very small grating to interval width ratio. Thus, we are actually using very thick gratings instead of thin gratings, which causes decrease in the illusory strength, as is discovered in \citep{soriano1996abutting}. Apart from the demonstrations given in Fig.\ref{Fig:AG-MNIST28}a, we can also alter more parameter settings, such as the direction of the gratings, yet we 
discovered that both vertical gratings or diagonal gratings are very difficult to recognize compared to the horizontal gratings, according to section 4.5. Generally, they also suffer from the low resolution of the original images. It is also worth noting that when the interval is set to 2, the pattern is not the standard abutting grating illusion anymore. The gratings are connected to each other at the corner diagonally, which makes them the abutting sinewave gratings, which is usually used as a comparison to abutting grating illusion in physiological studies\citep{song2006neural}\citep{grosof1993macaque}.

To test the robustness of deep learning models on AG-MNIST, we first train the models on the original training set of MNIST, then test them on both the original test set and the AG corrupted test set. We use the horizontal gratings with interval set to 2,4,6 and 8 respectively. The deep learning models we tested are fully connected neural network(FC) and convolutional neural network(CNN), both of which are trained and tested for 100 epochs. 

\subsection{Abutting grating illusion with high-resolution MNIST}
The original MNIST image samples have very low resolution, which does not allow testing deep learning models in a more convincing way. In order to produce high resolution number figures with stronger visual illusions, we can upsample the MNIST image samples by interpolating the images first, then apply the abutting grating corruption on the high-resolution images, as is depicted in Fig.\ref{Fig:abutting}c. We interpolated the original 28x28 images to 224x224 images, which is suitable for deep learning models trained with ImageNet dataset. Samples of high resolution number 8 are given in Fig.\ref{Fig:AG-MNIST28}b. It can be vividly seen that with a high-resolution image sample, we can preserve more details of the original image with the abutting grating corruption, leading to a much stronger illusory strength and more straightforward recognition compared to the low resolution counterparts. We depict not only the samples with horizontal gratings, but also vertical and diagonal gratings in Fig.\ref{Fig:AG-MNIST28}b. Similar to the low-resolution image samples, larger intervals leads to more difficulties to recognize the number, while small intervals induce much stronger visual illusions, which is consistent with the discovery in \citep{soriano1996abutting}.

Here we test four models, AlexNet, VGG11 with batch normalization, ResNet18, and DenseNet121 on the high resolution AG-MNIST. The model code was adapted from the torchvision library\citep{marcel2010torchvision}. In order to use these models on the MNIST dataset, we change the output of the last layer to 10 neurons. Moreover, we not only test the models on the horizontal gratings, but also vertical gratings and diagonal gratings. All four models are trained on the interpolated training set as well as tested on both high resolution MNIST and high resolution AG-MNIST for 100 epochs.
\subsection{Silhouettes of more complex objects(AG-silhouettes)}
Models trained on number figures are certainly not representative enough to measure the potential of deep learning models. While it is easy to create abutting grating illusion with MNIST, the task itself is not complex enough for us to fully understand the potential of deep learning models. A plethora of deep learning models have been developed and trained to recognize complex objects in natural images and have achieved almost equivalent complex object recognition performance as humans do. Thus it is a more convincing challenge to explore their ability of perceiving illusion of complex objects than simple number figures. However, generating abutting grating illusion with images of complex objects is more difficult compared to sheer number figures. Natural images contains abundant texture and local edge information, while we only need the outer contour or the silhouette to create the abutting grating illusions. Datasets such as Cifar10/100 or ImageNet are both not suitable to use directly, because drawing the outer contour is not a trivial task itself. 

A much easier choice is to use datasets that already provide the silhouette images. In their work to study the shape bias of deep learning models, \citep{geirhos2018imagenet} provided 160 images of objects without background. Later they converted the images to silhouette and edge versions and tested these samples with both humans and deep learning models. These silhouettes can be converted to abutting grating images following the same procedure of converting MNIST images. We produce the abutting grating illusion version of 160 silhouettes using horizontal, vertical, upper left to lower right and upper right to lower left gratings of interval from 4 to 14 with step 2, which in total would be 24 tests. In \citep{geirhos2018imagenet} a mapping function is provided to map 1,000 classes to 16 classes according to WordNet\citep{miller1995wordnet}. Samples are demonstrated in Fig.\ref{Fig:AG-MNIST28}c. By testing with these 160 images, we can test the illusory contour perception of various ImageNet pretrained models directly. 

Various genres of pretrained models are tested. We tested models of different architectures, from the classic architectures such as AlexNet, VggNet and ResNet, to modern architectures such as ViT and ConvNeXt. We tested all 63 pretrained models provided in torchvision 0.12 package and 22 pretrained models from timm 0.5.4 package. The model names are listed in Table \ref{tab:torchvision_models} and Table \ref{tab:timm_models}.

\begin{table}[]
\centering
\resizebox{\textwidth}{!}{%
\begin{tabular}{|l|l|}
\hline
AlexNet           & alexnet                                                       \\ \hline
VGG               & vgg11, vgg13, vgg16, vgg19                                    \\ \hline
VGG\_BN           & vgg11\_bn, vgg13\_bn, vgg16\_bn, vgg19\_bn                    \\ \hline
ResNet            & resnet18, resnet34, resnet50, resnet101, resnet152            \\ \hline
SqueezeNet        & squeezenet1\_0, squeezenet1\_1                                \\ \hline
DenseNet          & densenet121, densenet161, densenet169, densenet201            \\ \hline
Inception v3      & inception\_v3                                                 \\ \hline
GoogLeNet         & googlenet                                                     \\ \hline
ShuffleNet v2     & shufflenet\_v2\_x0\_5,shufflenet\_v2\_x1\_0                   \\ \hline
MobileNetV2       & mobilenet\_v2                                                 \\ \hline
MobileNetV3       & mobilenet\_v3\_large,mobilenet\_v3\_small                     \\ \hline
ResNeXt           & resnext50\_32x4d, resnext101\_32x8d                           \\ \hline
Wide ResNet       & wide\_resnet50\_2,wide\_resnet101\_2                          \\ \hline
MNASNet           & mnasnet0\_5,mnasnet1\_0                                       \\ \hline
EfficientNet &
  efficientnet\_b0,efficientnet\_b1,efficientnet\_b2,efficientnet\_b3,efficientnet\_b4,efficientnet\_b5,efficientnet\_b6,efficientnet\_b7 \\ \hline
RegNet &
  regnet\_y\_400mf,regnet\_y\_800mf,regnet\_y\_1\_6gf,regnet\_y\_3\_2gf,regnet\_y\_8gf,regnet\_y\_16gf,regnet\_y\_32gf,\\&regnet\_x\_400mf,regnet\_x\_800mf,regnet\_x\_1\_6gf,regnet\_x\_3\_2gf,regnet\_x\_8gf,regnet\_x\_16gf,regnet\_x\_32gf \\ \hline
VisionTransformer & vit\_b\_16,vit\_b\_32,vit\_l\_16,vit\_l\_32                   \\ \hline
ConvNeXt          & convnext\_tiny,convnext\_small,convnext\_base,convnext\_large \\ \hline
\end{tabular}%
}
\caption{Torchvision models tested on silhouettes and AG-silhouettes}
\label{tab:torchvision_models}
\end{table}

\begin{table}[]
\centering
\resizebox{\textwidth}{!}{%
\begin{tabular}{|l|l|}
\hline
MLPMixer        & mixer\_b16\_224,mixer\_l16\_224                                                                                                       \\ \hline
SwinTransformer & swin\_base\_patch4\_window7\_224,swin\_large\_patch4\_window7\_224,swin\_small\_patch4\_window7\_224,swin\_tiny\_patch4\_window7\_224 \\ \hline
Inception       & adv\_inception\_v3,ens\_adv\_inception\_resnet\_v2,gluon\_inception\_v3,inception\_resnet\_v2,inception\_v3,inception\_v4             \\ \hline
ConvNeXt &
  convnext\_base,convnext\_base\_384\_in22ft1k,convnext\_base\_in22ft1k,convnext\_large,convnext\_large\_384\_in22ft1k,\\&convnext\_large\_in22ft1k,convnext\_small,convnext\_tiny,convnext\_xlarge\_384\_in22ft1k,convnext\_xlarge\_in22ft1k \\ \hline
\end{tabular}%
}
\caption{Timm models tested on silhouettes and AG-silhouettes}
\label{tab:timm_models}
\end{table}

\begin{table}[]
\centering
\resizebox{\textwidth}{!}{%
\begin{tabular}{|l|l|}
\hline
Speckle Noise  &    resnet50\_speckle                                                                                            \\ \hline
Gaussian Noise & Gauss\_mult,Gauss\_sigma\_0.5                                                                  \\ \hline
SIN &
  alexnet\_trained\_on\_SIN,vgg16\_trained\_on\_SIN,\\&resnet50\_trained\_on\_SIN,resnet50\_trained\_on\_SIN\_and\_IN,resnet50\_trained\_on\_SIN\_and\_IN\_then\_finetuned\_on\_IN \\ \hline
ANT            & ANT,ANT\_SIN,ANT3x3,ANT3x3\_SIN                                                                \\ \hline
CutMix         & resnet50\_cutmix,resnet50\_feature\_cutmix,resnet101\_cutmix,resnet152\_cutmix,resnext\_cutmix \\ \hline
CutOut         & resnet50\_cutout                                                                               \\ \hline
mixup          & resnet50\_mixup,resnet50\_manifold\_mixup                                                      \\ \hline
AugMix         & resnet50\_augmix                                                                               \\ \hline
DeepAugment    & resnet50\_deepaugment,resnet50\_deepaugment\_augmix,resnext101\_32x8d\_deepaugment\_augmix     \\ \hline
\end{tabular}%
}
\caption{Data augmentation models tested on silhouettes and AG-silhouettes}
\label{tab:augmentation}
\end{table}
Apart from that, we also tested the influence of different data augmentation techniques. Various data augmentation techniques have been proposed to improve the robustness of deep learning models against unseen corruptions. In this experiment, we tested 24 models pretrained with ImageNet21k/ImageNet22k\citep{kolesnikov2020big}\citep{ridnik2021imagenet}\citep{codreanu2017scale}, Stylized ImageNet(SIN)\citep{geirhos2018imagenet}, speckle noise and gaussian noise, adversarial training\citep{wong2020fast}, adversarial noise training, Cutout\citep{devries2017improved}, MixUp\citep{zhang2017mixup}, Manifold MixUp\citep{verma2019manifold}, CutMix\citep{yun2019cutmix}, Augmix\citep{hendrycks2019augmix}, DeepAugment\citep{hendrycks2021many} and ConvNeXt. We provide a list of models we use in Table \ref{tab:augmentation}. All model weights are publicly available on github.

\subsection{Evaluating human performance}

On the contrary to the clean datasets, recognizing abutting grating corrupted images is a non-trivial task even for humans. Applying the AG corruption with large grating interval would eventually impair human perception. Thus, before we can trustfully test model generalization on illusory contours with the abutting grating corruption, it is crucial to determine the human performance under different conditions of abutting grating corruptions. 

We randomly sampled 100 images from the MNIST dataset, 10 for each number, and convert them to AG-MNIST with interval 4 and 6 of four grating directions, which makes 8 conditions in total. Another 
distinct 100 images are also sampled and converted to high-resolution AG-MNIST images with interval 4 and 8 of four grating directions, which makes 8 conditions as well. The same procedure and parameter settings of generating high-resolution AG-MNIST is also applied to 160 silhouette images. For all three datasets we tested 8 conditions, and for each condition, 3 subjects are tested. In total, 24 human subjects participated in our experiments, including 15 male subjects and 9 female subjects. We confirm that informed consent was obtained from all subjects. Most of the subjects are our colleagues who major in brain-inspired intelligence. 
None of them have been trained to recognize the abutting grating corrupted images.
Each subject is first presented with 100 AG-MNIST images, followed with 100 high-resolution AG-MNIST images, then 160 AG-silhouette images. Each subject is requested to classify the test images with no time limit for each image sample. The sizes of corrupted MNIST samples are approximately 0.7cm x 0.7cm on the screen, and the sizes of both corrupted high-resolution MNIST samples and  corrupted silhouette samples are approximately 5.6cm x 5.6cm. The image samples are presented at the approximate height of human eyes, and the distance between subject eyes and the screen is about 60cm. The subjects are allowed to move their heads around when recognizing the images, but the image sizes are always fixed.

\section{Results}
\subsection{Robustness on AG-MNIST}

\begin{figure}[htbp]
\centering
\includegraphics[scale=1]{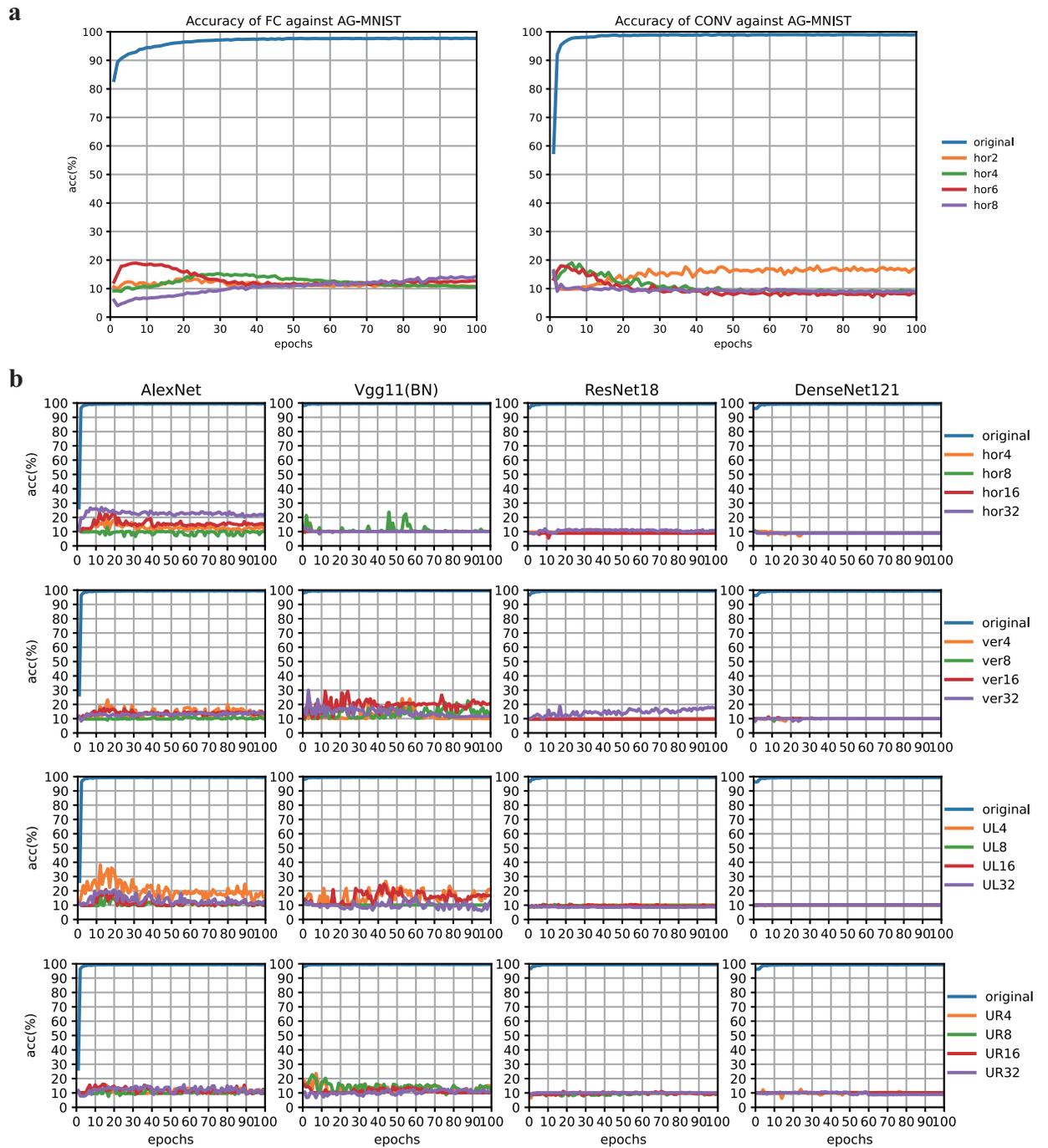}
\caption{\textbf{a}, Training FC and CNN for 100 epochs and testing on AG-MNIST with horizontal gratings of interval 2,4,6 and 8. \textbf{b}, Training AlexNet, Vgg11(BN), ResNet18 and DenseNet121 for 100 epochs and testing on high-resolution AG-MNIST with horizontal, vertical, upper left to lower right(UL) and upper right to lower left(UR) gratings. The interval is set to 4,8,16,32 respectively.}
\label{Fig:acc-mnist}
\end{figure}

The results are depicted in Fig.\ref{Fig:acc-mnist}a. While the results on the original test set grow fast and plateau at a very high accuracy for both models, the results on AG corrupted MNIST test set is much lower, with the best performance during the whole training procedure lower than 20\%. To be more specific, for the FC network, the result of interval 6 increases 
sharply at first, then gradually decays to a low level about a little higher than 10\%. The initial growth seems to be related to the growth of original test set curve, which also increases a lot at the first several epochs. The results of other three settings experience a much slower increase at first, then converge at approximately between 10\% and 15\%. On the other hand, for the CNN model, both curves of interval 4 and 6 have similar trend as the curve of interval 6 for FC model, both of which eventually converge to about the random level. The curve of interval 8 have a good start but then quickly drops to 10\% of accuracy. The curve of interval 2 is much more different from the others as it drops at first then gradually increases and plateaus at a relatively high accuracy compared to other curves. In general, we can see that the models seem to develop the ability of recognizing these AG transformed images at the start, but that ability quickly perished during the following training procedure, and remain at a extreme low level compared to the performance on the original test set.

\subsection{Robustness on high resolution AG-MNIST}
Fig.\ref{Fig:acc-mnist}b shows the performance of four models under four directions of gratings. We can see that the DenseNet121 generally performs the worst of all four models, as the curves maintain at the level of randomness during the whole training procedure, although there are also several small bumps. The ResNet18 is the second worst, randomly guessing in all occasions except for vertical gratings with interval 32, where the curve slowly grows to almost 20\% of accuracy. 
Compared to DenseNet121 and ResNet18, the performance of Vgg11(BN) and AlexNet is obviously better. The performance of Vgg11(BN) reaches higher than 20\% in several cases, and even surpasses 30\% upon vertical gratings of interval 32. The AlexNet performs even better, achieving almost 40\% of accuracy with the upper right gratings of interval 4. Nevertheless, although the curves oscillate in the first several epochs and reach a relatively high level, they all converge to a low accuracy. 

\subsection{Robustness on AG-silhouettes}

\begin{figure}[htbp]
\centering
\includegraphics[scale=0.42]{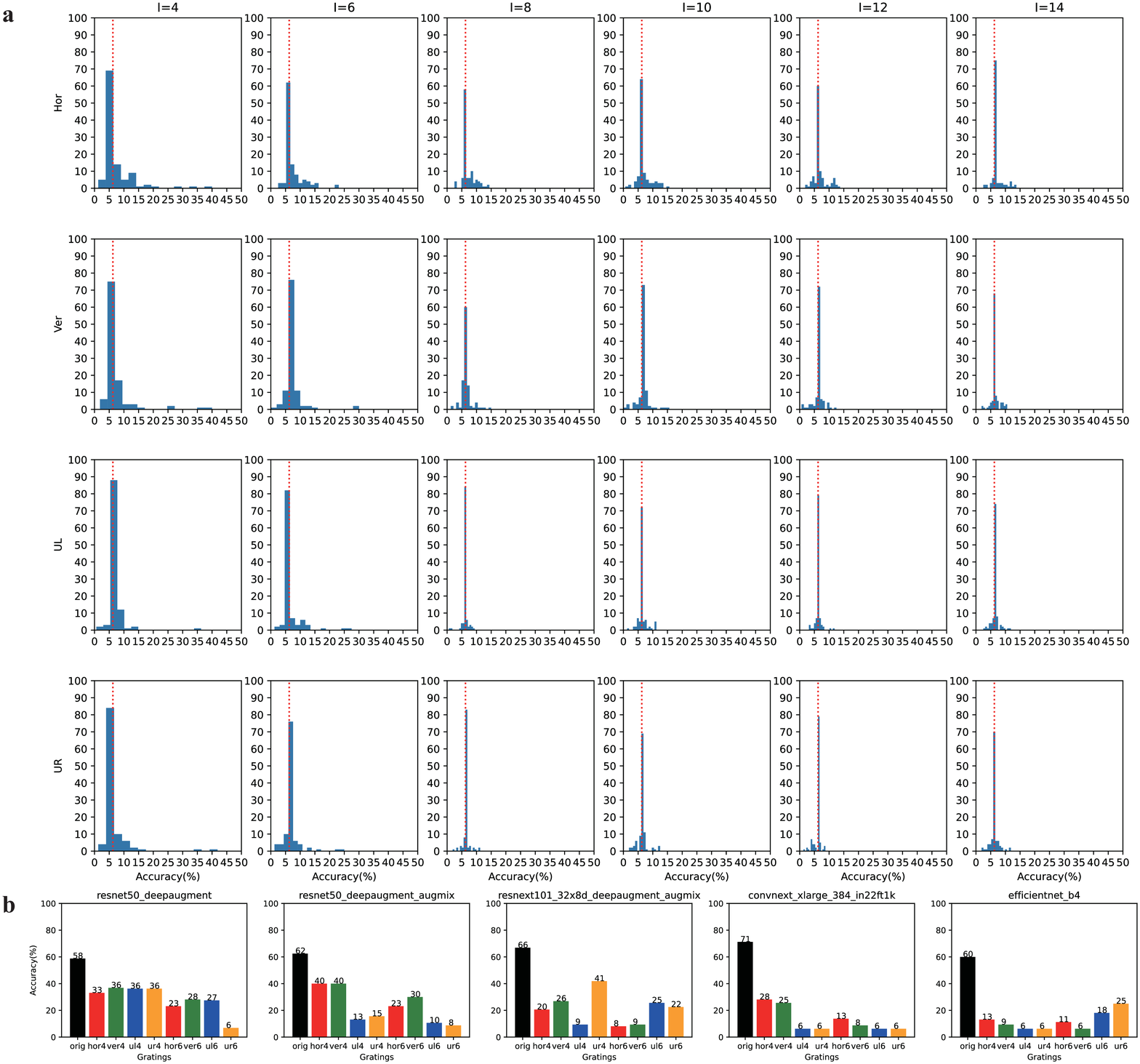}
\caption{\textbf{a}, The distribution of accuracy of 109 models under horizontal(Hor), vertical(Ver), upper left to lower right(UL), and upper right to lower left(UR) gratings. The interval is set to 4,6,8,10,12,14 respectively. The red dotted line represents the level of random guess, which is 6.25\%. \textbf{b}, five models that achieve results higher than 20\% under any condition. The performance on the original silhouettes, horizontal, vertical, upper left to lower right, upper right to lower left gratings of interval 4 and 6 are reported.}
\label{Fig:test_all_models}
\end{figure}
For each parameter setting of gratings, we plot the distributions of 109 model performance in Fig.\ref{Fig:test_all_models}a. For all 24 conditions, most results fall at or around the random guessing level. It can be clearly seen that the high blue columns generally coincide with the red dotted lines. There are also results fall at the nearby of the high blue columns. In fact, most ImageNet pretrained models provided in torchvision package as well as timm package perform almost randomly on the abutting grating illusion, including different variations of resnet, vision transformer, swin transformer. Except for the distributions around the random level, we also noticed that there are few outliers as well as big gaps between the outliers and the distributions when the interval settings is 4 or 6. These outliers indicate a significant improvement on the robustness against abutting grating illusions. In Fig.\ref{Fig:test_all_models}b, We report the performance of the corresponding models of these outliers by threshold 20\% of accuracy. Five models are reported in total whose performance are significantly better than the other models. Of all five models, three models are trained with DeepAugment technique, while the other two are ConvNeXt model and EfficientNet v4 respectively. Furthermore, we can see that the direction of gratings has a big impact on the model performance, as most models can only perform well on one direction setting and fail on the others. The resnet50 model trained solely with DeepAugment technique is more robust than other models against different directions, as it achieved good results almost under 7 conditions. Introducing the AugMix technique improves the resnet50 performance on several conditions but also causes big drop on others. Except for DeepAugment models, we also found that the ConvNeXt model performed relatively well compared to all other models, although still much worse than DeepAugment models. It is surprising since ConvNeXt has been one of the state-of-the-art models which performs much better than DeepAugment on various robustness benchmarks such as ImageNet-R\citep{hendrycks2021many}, ImageNet-C\citep{hooker2019compressed} and ImageNet-A\citep{hendrycks2021natural}. In fact, the performance of ConvNext model on the original silhouettes is about 71\% which is quite close to the average human performance of 75\%\citep{geirhos2018imagenet}. The performance of DeepAugment models on silhouettes images is worse, yet their performance on abutting grating illusions is better than ConvNeXt. Thus, the DeepAugment models experience a much smaller accuracy loss than ConvNeXt when the silhouette images are corrupted to their abutting grating versions.

\subsection{Explaining what models see}

\begin{figure}[htbp]
\centering
\includegraphics[scale=0.10]{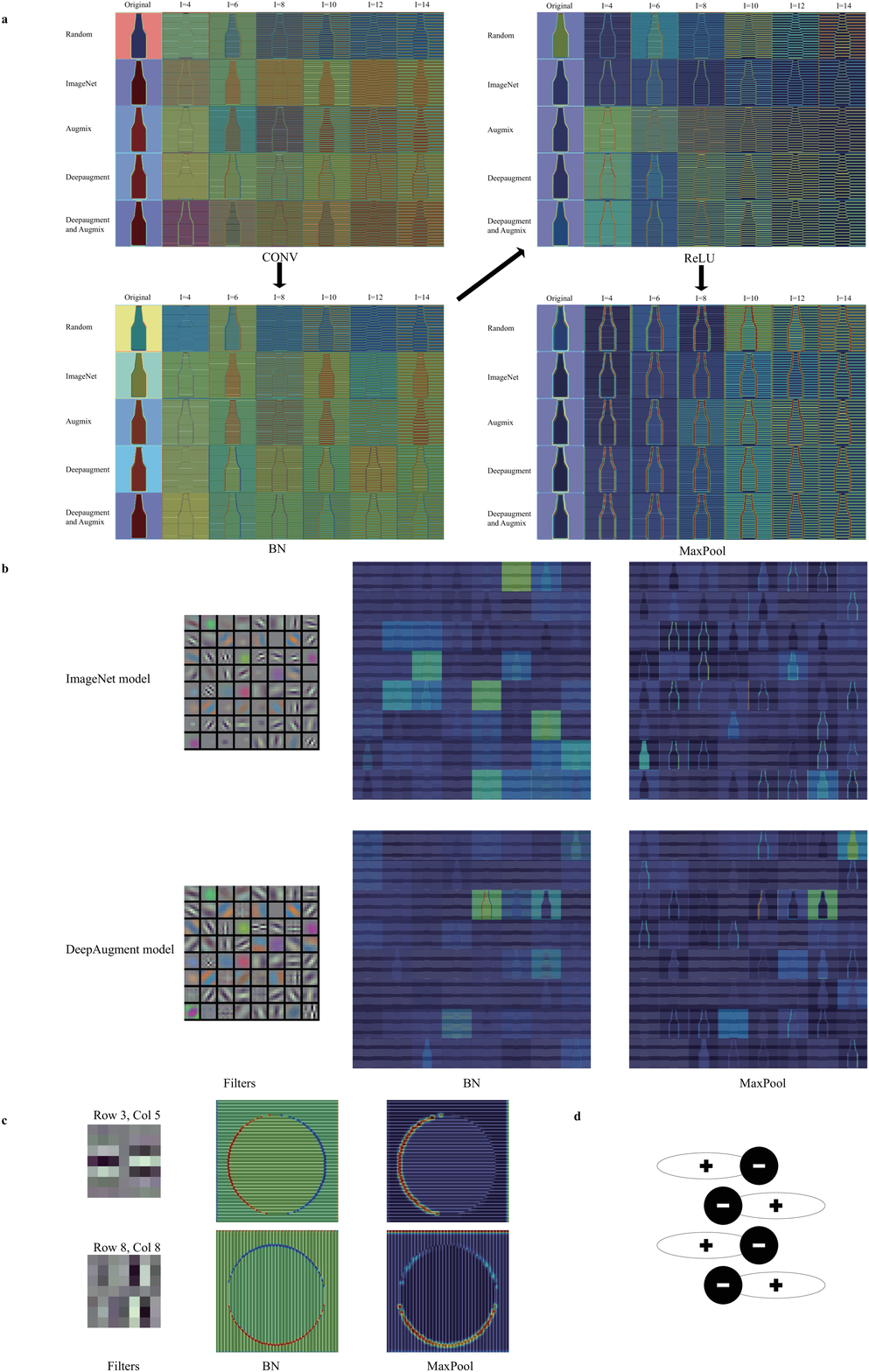}
\caption{\textbf{a}, The average activation maps for the early layers of ResNet50 models trained with different strategies. The activation maps are calculated for each convolutional filters and averaged together. It is later normalized to between 0 and 1. \textbf{b}, Visualization of first convolution layer filters and their corresponding  activation maps. The activation maps of all filters are normalized globally to between 0 and 1 for each filter. \textbf{c}, The filter of row 3, column 5 and the filter of row 8, column 8 of the DeepAugment model convolution filter map, their responses of the batch normalization layer and MaxPool layer. \textbf{d}, Hypothetical topology of end-stopped cells\citep{peterhans1986neuronal}.}
\label{Fig:Visualize}
\end{figure}

To understand why DeepAugment technique leads to such prominent improvement upon the recognition of abutting grating illusions, we visualise the early layers of the resnet50 model trained with DeepAugment as well as resnet50 trained with both DeepAugment and AugMix by computing the average activation map across the feature maps, which are normalized between 0 and 1 afterwards. In comparison, we also visualise the resnet50 models trained without DeepAugment, including an untrained resnet50 model with randomly initiated weights, resnet50 pretrained with ImageNet, resnet50 trained with augmix. We only visualise the early layers of the resnet50 models, 
including the first convolution layer, the batch normalization, the ReLU activation and the MaxPool layer. Early layers enables much higher resolutions than deeper layers, allowing exploring precise spatial phenomena as in neurophysiological studies of illusory contours.  Furthermore, we do not use explaining tools such as GradCAM\citep{selvaraju2017grad} or integrated gradients\citep{sundararajan2017axiomatic}. GradCAM is more targeted 
to show the significance of general areas in deeper layers, while the integrated gradients focuses on the input images, both not suitable for early layer explorations. Also, we found that applying the two techniques produced no meaningful patterns whatsoever. 
The average activation maps of four early layers are depicted in Fig.\ref{Fig:Visualize}a. The batch normalization layer demonstrates the most intriguing and clear pattern, where the models trained with DeepAugment as well as DeepAugment and AugMix exhibit strong activation on the one end of the gratings and depression on the other. This behaviour is quite similar to the end-stopping receptive field first discovered in \citep{hubel1965receptive}. It is found that the end-stopped cells (hypercomplex cells) are activated maximumly when the line-ends or corners are placed at the center of the receptive field, while elongating the line across the receptive field would reduce the activation. It has been widely believed to involve in the early processing of various illusory contour generation in biological systems, and multiple computational models are built upon it\citep{peterhans1986neuronal}\citep{finkel1989integration}\citep{lesher1995illusory}\citep{peterhans1989mechanisms}\citep{von1989mechanisms}\citep{heitger1994computational}\citep{francis1996cortical}.

When the grating interval equals 6,10 and 14, the activated ends of all gratings are on the same side, which elicits the phenomenon of strong activation along the left bottle edge. Likewise, strong depression can also be seen along the right edge. On the contrary, when the interval equals 4,8 or 12, the activation and depression ends alternate, which causes visually weaker responses along both sides. For models trained without DeepAugment, strangely, only the random model exhibits a similar yet reversed pattern when the interval is set to 6. Except for that, we cannot observe a clear activation or depression along the illusory bottle edges. In fact, in some cases, the grating ends are activated or depressed for models trained without DeepAugment. However, these models also exhibit strong activation or depression along the gratings themselves, which eventually weaken the pattern. Models trained with DeepAugment have suppressed either strong activation or depression along the gratings, which ultimately gives an evident perception of red and blue edges along the illusory bottle edges. Similar phenomenon can also be seen in the average activation map of the first convolution layer. However, only the model trained with DeepAugment alone demonstrates this phenomenon, and the response is not as strong as in batch normalization layer.

The average activation maps of ReLU layer and MaxPool layer have distinct patterns from the former two layers. The ReLU operation has turned both activation and depression into activation only, thus we can see clear activation along edges of both sides, especially for the map of MaxPool layer. More importantly, in the batch normalization layer, visually the complete illusory boundaries are not activated or depressed, as the space between the grating ends is not as activated or depressed as the end points. However, in the activation map of the ReLU layer, the space between the grating ends are also activated, especially when the grating interval is small, although not as strong as the activation of the end points. As the interval increases, the activation of middle space fades. The phenomenon is even stronger for the activation map of MaxPool. The MaxPool operation, which downgrades the input resolution, causes even activation along the illusory edges, where the activation of middle space are almost as strong as end points for all five models. Even increasing the grating intervals does not disconnect the activation between the end points. For interval 14, the models trained with DeepAugment seems to elicit stronger connections between end points than those without DeepAugment, but the difference is not significant enough.

We further visualize the globally normalized responses of each convolutional filter on the image of grating interval 6 in Fig.\ref{Fig:Visualize}b. We found that the activations of ImageNet model in BN layer is more dispersed than that of DeepAugment model. In addition, we discovered one specific filter(row 3, col 5) that have produced strongest response than all other filters. The response pattern obviously dominates the overall average activation map pattern. Comparing the activation map of this filter and the average activation map, we can see that the single filter has produced activation and depression more clear and continuous than the average activation map. Except for the BN layer, we also visualise the MaxPool layer. We can see that the filter produces strong and continuous activation along one side of the illusory bottle edges. In the average activation map of MaxPool, we found that all five models exhibit activation along the illusory edge. However, now we can see that the average illusory activation is the consequence of averaging across all filters for models trained without DeepAugment. On the other hand, for the model trained with DeepAugment, the right side of average illusory activation emerges from the average of all filters, while the left side is more likely to originate from one specific filter. Apart from the horizontal gratings, we also tested vertical gratings, and found that the filter at row 8, col 8 of DeepAugment feature map has similar behaviour as the filter at row 3, col 5. The diagonal gratings were tested as well, yet no meaningful results were found. We plot both filters and their corresponding responses at the BN layer and MaxPool layer in Fig.\ref{Fig:Visualize}c. Here we use a circle to demonstrate how illusory edges of different directions react to horizontal and vertical gratings.  We can see that an end-stopping property could be observed when the end-points are far from each other. While the end-points are getting close, their activation connects to each other, producing the illusory contour activation. Moreover, for the filter at row 3, col 5, generally we can see that the activation tend to be stronger when the gratings are more orthogonal to the illusory edges, which is consistent with the psychology experiment\citep{soriano1996abutting}. Intriguingly, these two filters resemble the hypothetical topology of end-stopped cells\citep{peterhans1986neuronal}\citep{lesher1995illusory}, which is depicted in Fig.\ref{Fig:Visualize}d.

\subsection{Evaluation of human performance}

\begin{figure}[htbp]
\centering
\includegraphics[scale=0.7]{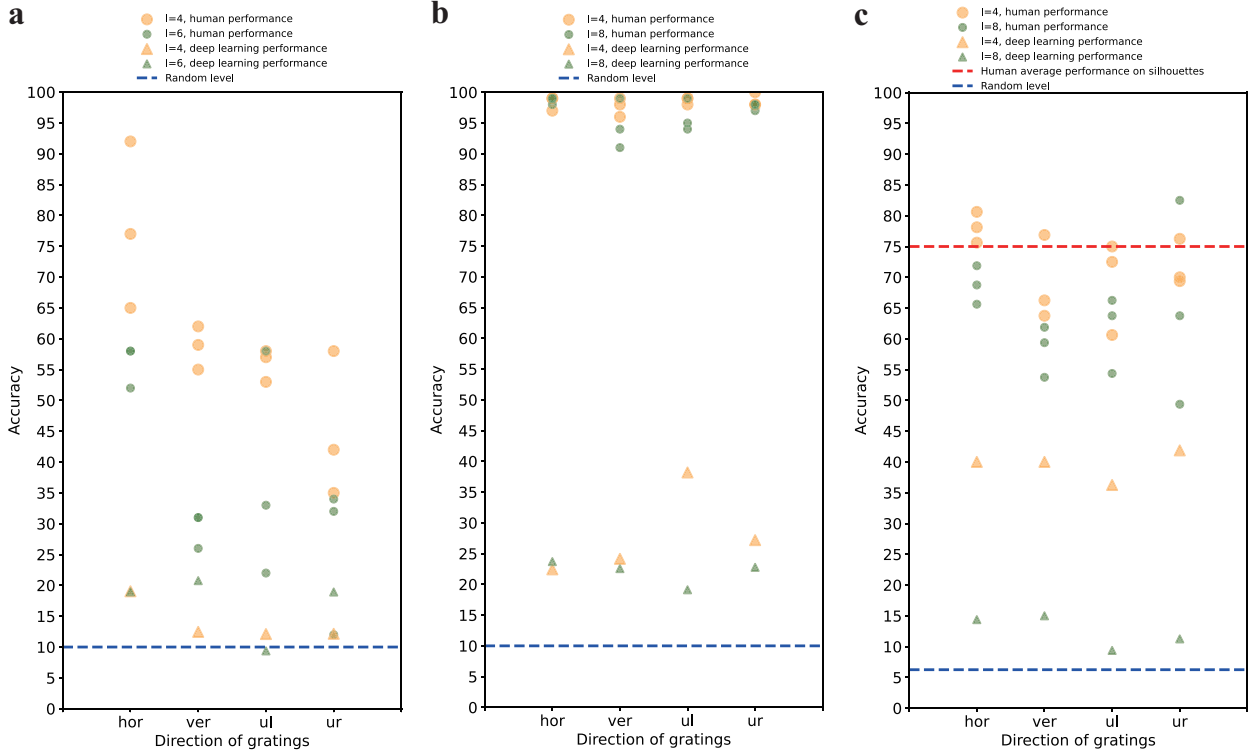}
\caption{\textbf{a}, Human performance on abutting grating images of 100 randomly selected samples of MNIST dataset, the grating interval set to 4 and 6 respectively. The best performance on the complete corrupted test set of deep learning models from section 4.1 is given as a comparison. \textbf{b}, Human performance on abutting grating images of 100 randomly selected samples of high-resolution MNIST dataset, the grating interval set to 4 and 8 respectively. The best performance on the complete corrupted test set of deep learning models from section 4.2 is given as a comparison. \textbf{c}, Human performance on abutting grating images of 160 silhouettes, the grating interval set to 4 and 6 respectively. The best performance of deep learning models from section 4.3 is given as a comparison. }
\label{Fig:Human}
\end{figure}

The human performance is given in Fig.\ref{Fig:Human}, each circle representing the performance of one subject. Fig.\ref{Fig:Human}a shows the human performance on the abutting grating images of 100 randomly selected MNIST samples. We also provide the best performance of FC and CNN on the complete corrupted test set from section 4.1, including the performance on the vertical, upper left and upper right directions of gratings. 
It is evident that horizontal gratings of interval 4 is simply the easiest condition for humans in all 8 conditions, as even the worst result is better than other conditions. Of the other three grating directions of interval 4, the performance on the vertical gratings and upper left gratings are very close, with the vertical condition little better. While the best performance of upper right gratings is as well as the upper left, the worst performance is much lower than the other conditions. For interval 6, the best performance is achieved by both horizontal and upper left gratings, reaching 58/100, however, it is very clear that the performance of other subjects on upper left gratings is much lower than that of horizontal gratings. Thus, the condition of horizontal gratings still outperforms other directions in general. On the other hand, we also provide the performance of deep learning models on the complete corrupted test set as a guideline. The performance of deep learning models are significantly worse than humans in most situations, with few exceptions where both humans and deep learning models are performing badly. Moreover, in several cases, the model performance is very close to randomness. In general, we found that humans tend to perform well with horizontal gratings, especially with interval 4, and the variations between subjects could be large. It is probably due to different levels of familiarity on the original MNIST dataset and strategies to recognize the abutting grating images.

Fig.\ref{Fig:Human}b gives the results on the high-resolution MNIST images. The average human performance is much better than the original MNIST, with smaller variations. The best results reach above 98/100 for all conditions. It can be concluded that the interpolation has allowed more information to be preserved during the abutting grating corruption. The best performance of deep learning models on complete corrupted test set is also given, which is relatively better than Fig.\ref{Fig:Human}a. Nevertheless, the gap between humans and deep learning models is still huge. Fig.\ref{Fig:Human}c gives the human performance on the abutting grating corrupted silhouette images. We also give the best recognition results from section 4.3. Moreover, the average human performance on the clean silhouette images is given for comparison, which is 75\% according to \citep{geirhos2018imagenet}. Similar to AG-MNIST, the human performance with the horizontal gratings of interval 4 surpasses other conditions, because all three subjects achieved even higher results than the average level of clear silhouettes. Although other directions of interval 4 do not lead to as good result as horizontal gratings, their best results reach or even surpass the average level of clear silhouettes. For interval 6, the overall performance is worse than interval 4, yet the overall gap is generally small compared to that of AG-MNIST tests. On the other hand, the performance of deep learning models on interval 4 is also better than that of interval 6, as expected. However, the increasing of grating intervals on silhouettes leads to a larger gap of performance than not only human performance, but also the gaps of AG-MNIST and high-resolution AG-MNIST. While pretrained deep learning models have achieved consistent results over all four directions and higher than performance on AG-MNIST and high-resolution AG-MNIST with interval 4, the increasing of grating intervals causes a huge drop. On the contrary, the human performance are not affected that much.

\section{Discussion}

Illusory contour perception, a fundamental ability of biological visual systems, remains rarely studied in deep learning realm. One of the core reasons is the difficulty of collecting datasets for illusory contours. In this study, we proposed abutting grating corruption based on the abutting grating illusion to generate various illusory contour datasets systematically. Using the abutting grating corruption, we can destroy all the local edge information while inducing strong illusory global contour perception for humans. Compared to mathematical corruption methods in the literature, our corruption has more profound relationship with human cognitive phenomenon, and can provide a unique tool to explore the difference between biological visual systems and deep learning models as well as to improve the robustness of future deep learning models. 

We apply the abutting grating corruption on the MNIST dataset. Most corruptions in the literature are assumed to have no influence on human perception, yet we come to aware that the abutting grating corruption could be too strong to damage human perceptions. Thus, we did human experiments to determine the reliability of the corruption. Note that due to the limit number of participants, the experiments only meant to provide a reference, instead of a solid conclusion of average human perception ability on the abutting grating datasets. By providing human performance on the subset of AG-MNIST, we found that AG-MNIST is challenging even for humans. We recommend using AG-MNIST with horizontal gratings of interval 4 for testing deep learning models trained with MNIST dataset. In order to strengthen the illusion, we propose to interpolate the original MNIST samples to high resolution. Experiments show that this method can significantly improve human recognition accuracy even with larger grating intervals. Thus the high resolution AG-MNIST is a much more convincing test for abutting grating illusions. Sadly, the high resolution MNIST is not the convention of training deep learning models to recognize hand written digits. Nevertheless, the interpolation method might be applied on other datasets to preserve more information and generate stronger illusion when using the abutting grating corruption.

The AG-MNIST dataset can only be used on models trained with MNIST dataset. However, the majority of modern deep learning models are developed to recognize more complex objects in natural images. Large quantity of pretrained models with various architectures and training techniques are available publicly, all of which cannot be tested directly with the AG-MNIST or toy datasets formerly proposed in the literature. Thus, we propose to 
apply the abutting grating corruption on the 160 silhouettes of complex objects. The 160 silhouettes was originally designed to test shape bias in ImageNet-trained deep learning models and can be directly converted to test robustness against abutting grating illusion and illusory contours. To our best knowledge, it is the first large-scale quantitative measurement on illusory contour perception of publicly available successful deep learning models. We are able to determine how different architectures of deep learning models, from the classic AlexNet or ResNet to more recent and popular vision transformer or ConvNeXt, react to abutting grating illusions and illusory contours. Our results show that although the architectures of deep learning models have evolved for decades and have conquered many obstacles, illusory contour perception remains a challenging problem. There are still huge gaps between the best performance of deep learning models and humans, and most commonly used deep learning models almost have zero ability of recognizing illusory contours. We may also conclude that the illusory contour perception is not the direct consequence of natural image statistics, as models trained with ImageNet or its variations can hardly recognize illusory contours. Nevertheless, we do not deny the possibility that ImageNet dataset itself might lack certain statistics of natural image statistics. Multiple data augmentation techniques are tested as well, all of them claimed to be robust against different corruptions, sometimes even unseen types. Almost all data augmentations do not help with the illusory contour perception, except for the DeepAugment technique. By plotting the distribution of classification accuracy, we discovered that models trained with DeepAugment clearly demonstrate significant improvement of illusory contour perception ability compared to all other models in the literature. The DeepAugment technique involves corrupting training images with neural networks, while illusory contours are not directly mentioned. We have yet to know why DeepAugment can generate such high robustness against abutting grating corruption, while losing the race on other robustness benchmark datasets. A possible explanation is that currently popular robustness datasets hardly capture the robustness of illusory contours. 

Our only clue might be the visualisation of early model layers. Strangely, we found that the MaxPool layer of all five models, including the one with random weights, exhibit activation and depression along the illusory bottle edges when averaged across all convolution filters. It indicates that the illusory activation at the MaxPool layer is independent from precise synaptic weights and emerges from the statistics of filter responses. As the phenomenon does not lead to the illusory contour classification ability, we can hardly conclude that this activation along the illusory edges has the same mechanism as the illusory contour activation studied in neuroscience whatsoever. On the other hand, we discovered obvious difference in convolution layer and batch normalization layer. Not only have we found end-stopping property in the average activation maps of DeepAugment models, but we have also discovered certain filters that resemble the receptive field of end-stopped neurons and theoretical topology of end-stopped neurons. The most intriguing part is that the end-stopped receptive fields as well as the theoretical topology are generated without manually designation. To our best knowledge, it is the first time that end-stopped neuron receptive fields and related neuronal dynamics are discovered and reported in artificial neural networks. As the end-stopping property is a common component of theoretical generation of illusory contours, the responses produced by these specific filters could be promising candidates for the illusory contour activation.  

In fact, it is almost impossible to be certain whether models can actually "see" illusory contours. Previous computational neuroscience models might be able to produce similar neural behaviours and to mimic the neuronal dynamics, but it does not guarantee the cognitive perceiving ability. \citep{bravo1988cats} proposed that if cats performed accurately in a task where humans found possible with the perception of illusory contours, and failed where human failed to perceive the illusory contours, then we might conclude that cats can perceive illusory contours. The abutting grating corruption allows testing deep learning models in a similar way. Nevertheless, the perception of illusory contours should be considered in both psychophysiology and neurophysiology. Models are expected to exhibit similar neuronal behaviours as in neuroscience studies as well as to actually recognize the object by integrating the illusory information. The abutting grating corruption is designed for the latter. Luckily, the corruption helps find the DeepAugment model, which is the very first model in the literature that not only can perform cognitive tasks on both natural image objects classification and illusory object classification, but also exhibits the similar behaviour to the neuronal dynamics of illusory contours. We expect this model to be helpful in future computational neuroscience studies.

In general, we have constructed a cognitive based corruption method to systematically evaluate the perception ability of modern deep learning models on abutting grating illusion and illusory contour. The illusory contour perception should be an inevitable ability of future deep learning models due to its wide existence in biological visual systems. We hope that future deep learning models would be tested with our corruption. Also, we focused on the abutting grating illusion, but similar corruptions might be created to study other types of illusory contours or visual illusions. We have tested on 109 publicly available deep learning models while more models in the literature could be tested. In addition, we will further explore why DeepAugment improve the robustness so much and try to improve the results further. Moreover, we use simple visualization on early layers of the DeepAugment models. Deeper layers might be studied with more powerful visualization methods because neurophysiological studies showed that deeper layers of visual cortex also react to illusory contours. Last but not least, we conducted human experiments on the abutting grating datasets. Future work might involve a more thorough study of the humans limitation on different abutting grating datasets and diverse parameter settings.

\bibliographystyle{unsrtnat}
\bibliography{references}  

\begin{thebibliography}{86}
\providecommand{\natexlab}[1]{#1}
\providecommand{\url}[1]{\texttt{#1}}
\expandafter\ifx\csname urlstyle\endcsname\relax
  \providecommand{\doi}[1]{doi: #1}\else
  \providecommand{\doi}{doi: \begingroup \urlstyle{rm}\Url}\fi

\bibitem[Russakovsky et~al.(2015)Russakovsky, Deng, Su, Krause, Satheesh, Ma,
  Huang, Karpathy, Khosla, Bernstein, et~al.]{russakovsky2015imagenet}
Olga Russakovsky, Jia Deng, Hao Su, Jonathan Krause, Sanjeev Satheesh, Sean Ma,
  Zhiheng Huang, Andrej Karpathy, Aditya Khosla, Michael Bernstein, et~al.
\newblock Imagenet large scale visual recognition challenge.
\newblock \emph{International journal of computer vision}, 115\penalty0
  (3):\penalty0 211--252, 2015.

\bibitem[Dodge and Karam(2017)]{dodge2017study}
Samuel Dodge and Lina Karam.
\newblock A study and comparison of human and deep learning recognition
  performance under visual distortions.
\newblock In \emph{2017 26th international conference on computer communication
  and networks (ICCCN)}, pages 1--7. IEEE, 2017.

\bibitem[Dodge and Karam(2016)]{dodge2016understanding}
Samuel Dodge and Lina Karam.
\newblock Understanding how image quality affects deep neural networks.
\newblock In \emph{2016 eighth international conference on quality of
  multimedia experience (QoMEX)}, pages 1--6. IEEE, 2016.

\bibitem[Hendrycks and Dietterich(2019)]{hendrycks2019benchmarking}
Dan Hendrycks and Thomas Dietterich.
\newblock Benchmarking neural network robustness to common corruptions and
  perturbations.
\newblock \emph{arXiv preprint arXiv:1903.12261}, 2019.

\bibitem[Szegedy et~al.(2013)Szegedy, Zaremba, Sutskever, Bruna, Erhan,
  Goodfellow, and Fergus]{szegedy2013intriguing}
Christian Szegedy, Wojciech Zaremba, Ilya Sutskever, Joan Bruna, Dumitru Erhan,
  Ian Goodfellow, and Rob Fergus.
\newblock Intriguing properties of neural networks.
\newblock \emph{arXiv preprint arXiv:1312.6199}, 2013.

\bibitem[Carlini and Wagner(2017)]{carlini2017towards}
Nicholas Carlini and David Wagner.
\newblock Towards evaluating the robustness of neural networks.
\newblock In \emph{2017 ieee symposium on security and privacy (sp)}, pages
  39--57. IEEE, 2017.

\bibitem[Madry et~al.(2017)Madry, Makelov, Schmidt, Tsipras, and
  Vladu]{madry2017towards}
Aleksander Madry, Aleksandar Makelov, Ludwig Schmidt, Dimitris Tsipras, and
  Adrian Vladu.
\newblock Towards deep learning models resistant to adversarial attacks.
\newblock \emph{arXiv preprint arXiv:1706.06083}, 2017.

\bibitem[Moosavi-Dezfooli et~al.(2016)Moosavi-Dezfooli, Fawzi, and
  Frossard]{moosavi2016deepfool}
Seyed-Mohsen Moosavi-Dezfooli, Alhussein Fawzi, and Pascal Frossard.
\newblock Deepfool: a simple and accurate method to fool deep neural networks.
\newblock In \emph{Proceedings of the IEEE conference on computer vision and
  pattern recognition}, pages 2574--2582, 2016.

\bibitem[Papernot et~al.(2016)Papernot, McDaniel, Wu, Jha, and
  Swami]{papernot2016distillation}
Nicolas Papernot, Patrick McDaniel, Xi~Wu, Somesh Jha, and Ananthram Swami.
\newblock Distillation as a defense to adversarial perturbations against deep
  neural networks.
\newblock In \emph{2016 IEEE symposium on security and privacy (SP)}, pages
  582--597. IEEE, 2016.

\bibitem[Borji and Itti(2014)]{borji2014human}
Ali Borji and Laurent Itti.
\newblock Human vs. computer in scene and object recognition.
\newblock In \emph{Proceedings of the IEEE conference on computer vision and
  pattern recognition}, pages 113--120, 2014.

\bibitem[Fleuret et~al.(2011)Fleuret, Li, Dubout, Wampler, Yantis, and
  Geman]{fleuret2011comparing}
Fran{\c{c}}ois Fleuret, Ting Li, Charles Dubout, Emma~K Wampler, Steven Yantis,
  and Donald Geman.
\newblock Comparing machines and humans on a visual categorization test.
\newblock \emph{Proceedings of the National Academy of Sciences}, 108\penalty0
  (43):\penalty0 17621--17625, 2011.

\bibitem[Stabinger et~al.(2016)Stabinger, Rodr{\'\i}guez-S{\'a}nchez, and
  Piater]{stabinger201625}
Sebastian Stabinger, Antonio Rodr{\'\i}guez-S{\'a}nchez, and Justus Piater.
\newblock 25 years of cnns: Can we compare to human abstraction capabilities?
\newblock In \emph{International conference on artificial neural networks},
  pages 380--387. Springer, 2016.

\bibitem[Parikh(2011)]{parikh2011recognizing}
Devi Parikh.
\newblock Recognizing jumbled images: The role of local and global information
  in image classification.
\newblock In \emph{2011 International Conference on Computer Vision}, pages
  519--526. IEEE, 2011.

\bibitem[Hendrycks et~al.(2021{\natexlab{a}})Hendrycks, Basart, Mu, Kadavath,
  Wang, Dorundo, Desai, Zhu, Parajuli, Guo, et~al.]{hendrycks2021many}
Dan Hendrycks, Steven Basart, Norman Mu, Saurav Kadavath, Frank Wang, Evan
  Dorundo, Rahul Desai, Tyler Zhu, Samyak Parajuli, Mike Guo, et~al.
\newblock The many faces of robustness: A critical analysis of
  out-of-distribution generalization.
\newblock In \emph{Proceedings of the IEEE/CVF International Conference on
  Computer Vision}, pages 8340--8349, 2021{\natexlab{a}}.

\bibitem[Ford et~al.(2019)Ford, Gilmer, Carlini, and
  Cubuk]{ford2019adversarial}
Nic Ford, Justin Gilmer, Nicolas Carlini, and Dogus Cubuk.
\newblock Adversarial examples are a natural consequence of test error in
  noise.
\newblock \emph{arXiv preprint arXiv:1901.10513}, 2019.

\bibitem[Lopes et~al.(2019)Lopes, Yin, Poole, Gilmer, and
  Cubuk]{lopes2019improving}
Raphael~Gontijo Lopes, Dong Yin, Ben Poole, Justin Gilmer, and Ekin~D Cubuk.
\newblock Improving robustness without sacrificing accuracy with patch gaussian
  augmentation.
\newblock \emph{arXiv preprint arXiv:1906.02611}, 2019.

\bibitem[Hendrycks et~al.(2019{\natexlab{a}})Hendrycks, Mu, Cubuk, Zoph,
  Gilmer, and Lakshminarayanan]{hendrycks2019augmix}
Dan Hendrycks, Norman Mu, Ekin~D Cubuk, Barret Zoph, Justin Gilmer, and Balaji
  Lakshminarayanan.
\newblock Augmix: A simple data processing method to improve robustness and
  uncertainty.
\newblock \emph{arXiv preprint arXiv:1912.02781}, 2019{\natexlab{a}}.

\bibitem[Rusak et~al.(2020)Rusak, Schott, Zimmermann, Bitterwolfb, Bringmann,
  Bethge, and Brendel]{rusak2020increasing}
Evgenia Rusak, Lukas Schott, Roland Zimmermann, Julian Bitterwolfb, Oliver
  Bringmann, Matthias Bethge, and Wieland Brendel.
\newblock Increasing the robustness of dnns against im-age corruptions by
  playing the game of noise.
\newblock 2020.

\bibitem[Goodfellow et~al.(2014)Goodfellow, Shlens, and
  Szegedy]{goodfellow2014explaining}
Ian~J Goodfellow, Jonathon Shlens, and Christian Szegedy.
\newblock Explaining and harnessing adversarial examples.
\newblock \emph{arXiv preprint arXiv:1412.6572}, 2014.

\bibitem[Laidlaw et~al.(2020)Laidlaw, Singla, and Feizi]{laidlaw2020perceptual}
Cassidy Laidlaw, Sahil Singla, and Soheil Feizi.
\newblock Perceptual adversarial robustness: Defense against unseen threat
  models.
\newblock \emph{arXiv preprint arXiv:2006.12655}, 2020.

\bibitem[Hendrycks et~al.(2019{\natexlab{b}})Hendrycks, Lee, and
  Mazeika]{hendrycks2019using}
Dan Hendrycks, Kimin Lee, and Mantas Mazeika.
\newblock Using pre-training can improve model robustness and uncertainty.
\newblock In \emph{International Conference on Machine Learning}, pages
  2712--2721. PMLR, 2019{\natexlab{b}}.

\bibitem[Landau et~al.(1988)Landau, Smith, and Jones]{landau1988importance}
Barbara Landau, Linda~B Smith, and Susan~S Jones.
\newblock The importance of shape in early lexical learning.
\newblock \emph{Cognitive development}, 3\penalty0 (3):\penalty0 299--321,
  1988.

\bibitem[Gershkoff-Stowe and Smith(2004)]{gershkoff2004shape}
Lisa Gershkoff-Stowe and Linda~B Smith.
\newblock Shape and the first hundred nouns.
\newblock \emph{Child development}, 75\penalty0 (4):\penalty0 1098--1114, 2004.

\bibitem[Hosseini et~al.(2018)Hosseini, Xiao, Jaiswal, and
  Poovendran]{hosseini2018assessing}
Hossein Hosseini, Baicen Xiao, Mayoore Jaiswal, and Radha Poovendran.
\newblock Assessing shape bias property of convolutional neural networks.
\newblock In \emph{Proceedings of the IEEE Conference on Computer Vision and
  Pattern Recognition Workshops}, pages 1923--1931, 2018.

\bibitem[Gatys et~al.(2017)Gatys, Ecker, and Bethge]{gatys2017texture}
Leon~A Gatys, Alexander~S Ecker, and Matthias Bethge.
\newblock Texture and art with deep neural networks.
\newblock \emph{Current opinion in neurobiology}, 46:\penalty0 178--186, 2017.

\bibitem[Brendel and Bethge(2019)]{brendel2019approximating}
Wieland Brendel and Matthias Bethge.
\newblock Approximating cnns with bag-of-local-features models works
  surprisingly well on imagenet.
\newblock \emph{arXiv preprint arXiv:1904.00760}, 2019.

\bibitem[Geirhos et~al.(2018)Geirhos, Rubisch, Michaelis, Bethge, Wichmann, and
  Brendel]{geirhos2018imagenet}
Robert Geirhos, Patricia Rubisch, Claudio Michaelis, Matthias Bethge, Felix~A
  Wichmann, and Wieland Brendel.
\newblock Imagenet-trained cnns are biased towards texture; increasing shape
  bias improves accuracy and robustness.
\newblock \emph{arXiv preprint arXiv:1811.12231}, 2018.

\bibitem[Hermann et~al.(2020)Hermann, Chen, and Kornblith]{hermann2020origins}
Katherine Hermann, Ting Chen, and Simon Kornblith.
\newblock The origins and prevalence of texture bias in convolutional neural
  networks.
\newblock \emph{Advances in Neural Information Processing Systems},
  33:\penalty0 19000--19015, 2020.

\bibitem[Kubilius et~al.(2016)Kubilius, Bracci, and Op~de
  Beeck]{kubilius2016deep}
Jonas Kubilius, Stefania Bracci, and Hans~P Op~de Beeck.
\newblock Deep neural networks as a computational model for human shape
  sensitivity.
\newblock \emph{PLoS computational biology}, 12\penalty0 (4):\penalty0
  e1004896, 2016.

\bibitem[Ritter et~al.(2017)Ritter, Barrett, Santoro, and
  Botvinick]{ritter2017cognitive}
Samuel Ritter, David~GT Barrett, Adam Santoro, and Matt~M Botvinick.
\newblock Cognitive psychology for deep neural networks: A shape bias case
  study.
\newblock In \emph{International conference on machine learning}, pages
  2940--2949. PMLR, 2017.

\bibitem[Kanizsa(1974)]{kanizsa1974contours}
Gaetano Kanizsa.
\newblock Contours without gradients or cognitive contours?
\newblock \emph{Giornale Italiano di Psicologia}, 1974.

\bibitem[von~der Heydt and Peterhans(1989)]{von1989mechanisms}
Riidiger von~der Heydt and Esther Peterhans.
\newblock Mechanisms of contour perception in monkey visual cortex. i. lines of
  pattern discontinuity.
\newblock \emph{Journal of Neuroscience}, 9\penalty0 (5):\penalty0 1731--1748,
  1989.

\bibitem[Gurnsey et~al.(1992)Gurnsey, Humphrey, and
  Kapitan]{gurnsey1992parallel}
Rick Gurnsey, G~Keith Humphrey, and Paula Kapitan.
\newblock Parallel discrimination of subjective contours defined by offset
  gratings.
\newblock \emph{Perception \& Psychophysics}, 52\penalty0 (3):\penalty0
  263--276, 1992.

\bibitem[Davis and Driver(1998)]{davis1998kanizsa}
Greg Davis and Jon Driver.
\newblock Kanizsa subjective figures can act as occluding surfaces at parallel
  stages of visual search.
\newblock \emph{Journal of Experimental Psychology: Human Perception and
  Performance}, 24\penalty0 (1):\penalty0 169, 1998.

\bibitem[Schumann(1918)]{schumann1918beitrage}
Friedrich Schumann.
\newblock \emph{Beitr{\"a}ge zur analyse der gesichtswahrnehmungen}.
\newblock Number 4-7. JA Barth, 1918.

\bibitem[Canny(1986)]{canny1986computational}
John Canny.
\newblock A computational approach to edge detection.
\newblock \emph{IEEE Transactions on pattern analysis and machine
  intelligence}, \penalty0 (6):\penalty0 679--698, 1986.

\bibitem[Pan et~al.(2012)Pan, Chen, Yin, An, Zhang, Lu, Gong, Li, and
  Wang]{pan2012equivalent}
Yanxia Pan, Minggui Chen, Jiapeng Yin, Xu~An, Xian Zhang, Yiliang Lu, Hongliang
  Gong, Wu~Li, and Wei Wang.
\newblock Equivalent representation of real and illusory contours in macaque
  v4.
\newblock \emph{Journal of Neuroscience}, 32\penalty0 (20):\penalty0
  6760--6770, 2012.

\bibitem[De~Weerd et~al.(1996)De~Weerd, Desimone, and Ungerleider]{de1996cue}
Peter De~Weerd, Robert Desimone, and Leslie~G Ungerleider.
\newblock Cue-dependent deficits in grating orientation discrimination after v4
  lesions in macaques.
\newblock \emph{Visual neuroscience}, 13\penalty0 (3):\penalty0 529--538, 1996.

\bibitem[Montaser-Kouhsari et~al.(2007)Montaser-Kouhsari, Landy, Heeger, and
  Larsson]{montaser2007orientation}
Leila Montaser-Kouhsari, Michael~S Landy, David~J Heeger, and Jonas Larsson.
\newblock Orientation-selective adaptation to illusory contours in human visual
  cortex.
\newblock \emph{Journal of Neuroscience}, 27\penalty0 (9):\penalty0 2186--2195,
  2007.

\bibitem[Ramsden et~al.(2001)Ramsden, Hung, and Roe]{ramsden2001real}
Benjamin~M Ramsden, Chou~P Hung, and Anna~Wang Roe.
\newblock Real and illusory contour processing in area v1 of the primate: a
  cortical balancing act.
\newblock \emph{Cerebral Cortex}, 11\penalty0 (7):\penalty0 648--665, 2001.

\bibitem[Kanizsa(1976)]{kanizsa1976subjective}
Gaetano Kanizsa.
\newblock Subjective contours.
\newblock \emph{Scientific American}, 234\penalty0 (4):\penalty0 48--53, 1976.

\bibitem[Ehrenstein(1941)]{ehrenstein1941abwandlungen}
Walter Ehrenstein.
\newblock {\"U}ber abwandlungen der l. hermannschen helligkeitserscheinung.
\newblock \emph{Zeitschrift f{\"u}r Psychologie: Organ der Deutschen
  Gesellschaft f{\"u}r Psychologie}, 1941.

\bibitem[Baker et~al.(2018)Baker, Erlikhman, Kellman, and Lu]{baker2018deep}
Nicholas Baker, Gennady Erlikhman, Philip~J Kellman, and Hongjing Lu.
\newblock Deep convolutional networks do not perceive illusory contours.
\newblock In \emph{CogSci}, 2018.

\bibitem[Kellman et~al.(2017)Kellman, Baker, Erlikhman, and
  Lu]{kellman2017classification}
Philip Kellman, Nicholas Baker, Gennady Erlikhman, and Hongjing Lu.
\newblock Classification images reveal that deep learning networks fail to
  perceive illusory contours.
\newblock \emph{Journal of vision}, 17\penalty0 (10):\penalty0 569--569, 2017.

\bibitem[Ringach and Shapley(1996)]{ringach1996spatial}
Dario~L Ringach and Robert Shapley.
\newblock Spatial and temporal properties of illusory contours and amodal
  boundary completion.
\newblock \emph{Vision research}, 36\penalty0 (19):\penalty0 3037--3050, 1996.

\bibitem[Lotter et~al.(2018)Lotter, Kreiman, and Cox]{lotter2018neural}
William Lotter, Gabriel Kreiman, and David Cox.
\newblock A neural network trained to predict future video frames mimics
  critical properties of biological neuronal responses and perception.
\newblock \emph{arXiv preprint arXiv:1805.10734}, 2018.

\bibitem[Pang et~al.(2021)Pang, O’May, Choksi, and
  VanRullen]{pang2021predictive}
Zhaoyang Pang, Callum~Biggs O’May, Bhavin Choksi, and Rufin VanRullen.
\newblock Predictive coding feedback results in perceived illusory contours in
  a recurrent neural network.
\newblock \emph{Neural Networks}, 144:\penalty0 164--175, 2021.

\bibitem[Rao and Ballard(1999)]{rao1999predictive}
Rajesh~PN Rao and Dana~H Ballard.
\newblock Predictive coding in the visual cortex: a functional interpretation
  of some extra-classical receptive-field effects.
\newblock \emph{Nature neuroscience}, 2\penalty0 (1):\penalty0 79--87, 1999.

\bibitem[Hubel and Wiesel(1965)]{hubel1965receptive}
David~H Hubel and Torsten~N Wiesel.
\newblock Receptive fields and functional architecture in two nonstriate visual
  areas (18 and 19) of the cat.
\newblock \emph{Journal of neurophysiology}, 28\penalty0 (2):\penalty0
  229--289, 1965.

\bibitem[Hubel and Wiesel(1968)]{hubel1968receptive}
David~H Hubel and Torsten~N Wiesel.
\newblock Receptive fields and functional architecture of monkey striate
  cortex.
\newblock \emph{The Journal of physiology}, 195\penalty0 (1):\penalty0
  215--243, 1968.

\bibitem[Peterhans et~al.(1986)Peterhans, Von~der Heydt, and
  Baumgartner]{peterhans1986neuronal}
E~Peterhans, R~Von~der Heydt, and G~Baumgartner.
\newblock Neuronal responses to illusory contour stimuli reveal stages of
  visual cortical processing.
\newblock \emph{Visual neuroscience}, pages 343--351, 1986.

\bibitem[Finkel and Edelman(1989)]{finkel1989integration}
L~Hv Finkel and Gerald~M Edelman.
\newblock Integration of distributed cortical systems by reentry: A computer
  simulation of interactive functionally segregated visual areas.
\newblock \emph{Journal of Neuroscience}, 9\penalty0 (9):\penalty0 3188--3208,
  1989.

\bibitem[Lesher(1995)]{lesher1995illusory}
Gregory~W Lesher.
\newblock Illusory contours: Toward a neurally based perceptual theory.
\newblock \emph{Psychonomic Bulletin \& Review}, 2\penalty0 (3):\penalty0
  279--321, 1995.

\bibitem[Peterhans and von~der Heydt(1989)]{peterhans1989mechanisms}
Esther Peterhans and Riidiger von~der Heydt.
\newblock Mechanisms of contour perception in monkey visual cortex. ii.
  contours bridging gaps.
\newblock \emph{Journal of Neuroscience}, 9\penalty0 (5):\penalty0 1749--1763,
  1989.

\bibitem[Heitger et~al.(1994)Heitger, von~der Heydt, and
  Kubler]{heitger1994computational}
Friedrich Heitger, R{\"u}diger von~der Heydt, and Olaf Kubler.
\newblock A computational model of neural contour processing: Figure-ground
  segregation and illusory contours.
\newblock In \emph{Proceedings of PerAc'94. From Perception to Action}, pages
  181--192. IEEE, 1994.

\bibitem[Francis and Grossberg(1996)]{francis1996cortical}
Gregory Francis and Stephen Grossberg.
\newblock Cortical dynamics of form and motion integration: Persistence,
  apparent motion, and illusory contours.
\newblock \emph{Vision Research}, 36\penalty0 (1):\penalty0 149--173, 1996.

\bibitem[Nieder(2002)]{nieder2002seeing}
Andreas Nieder.
\newblock Seeing more than meets the eye: processing of illusory contours in
  animals.
\newblock \emph{Journal of Comparative Physiology A}, 188\penalty0
  (4):\penalty0 249--260, 2002.

\bibitem[Kanizsa(1955)]{kanizsa1955margini}
Gaetano Kanizsa.
\newblock Margini quasi-percettivi in campi con stimolazione omogenea.
\newblock \emph{Rivista di psicologia}, 49\penalty0 (1):\penalty0 7--30, 1955.

\bibitem[Lee and Nguyen(2001)]{lee2001dynamics}
Tai~Sing Lee and My~Nguyen.
\newblock Dynamics of subjective contour formation in the early visual cortex.
\newblock \emph{Proceedings of the National Academy of Sciences}, 98\penalty0
  (4):\penalty0 1907--1911, 2001.

\bibitem[Bakin et~al.(2000)Bakin, Nakayama, and Gilbert]{bakin2000visual}
Jonathan~S Bakin, Ken Nakayama, and Charles~D Gilbert.
\newblock Visual responses in monkey areas v1 and v2 to three-dimensional
  surface configurations.
\newblock \emph{Journal of Neuroscience}, 20\penalty0 (21):\penalty0
  8188--8198, 2000.

\bibitem[Cox et~al.(2013)Cox, Schmid, Peters, Saunders, Leopold, and
  Maier]{cox2013receptive}
Michele~A Cox, Michael~C Schmid, Andrew~J Peters, Richard~C Saunders, David~A
  Leopold, and Alexander Maier.
\newblock Receptive field focus of visual area v4 neurons determines responses
  to illusory surfaces.
\newblock \emph{Proceedings of the National Academy of Sciences}, 110\penalty0
  (42):\penalty0 17095--17100, 2013.

\bibitem[Von~der Heydt et~al.(1984)Von~der Heydt, Peterhans, and
  Baumgartner]{von1984illusory}
R{\"u}diger Von~der Heydt, Esther Peterhans, and Gunter Baumgartner.
\newblock Illusory contours and cortical neuron responses.
\newblock \emph{Science}, 224\penalty0 (4654):\penalty0 1260--1262, 1984.

\bibitem[Grosof et~al.(1993)Grosof, Shapley, and Hawken]{grosof1993macaque}
David~H Grosof, Robert~M Shapley, and Michael~J Hawken.
\newblock Macaque vi neurons can signal ‘illusory’contours.
\newblock \emph{Nature}, 365\penalty0 (6446):\penalty0 550--552, 1993.

\bibitem[Gatys et~al.(2016)Gatys, Ecker, and Bethge]{gatys2016image}
Leon~A Gatys, Alexander~S Ecker, and Matthias Bethge.
\newblock Image style transfer using convolutional neural networks.
\newblock In \emph{Proceedings of the IEEE conference on computer vision and
  pattern recognition}, pages 2414--2423, 2016.

\bibitem[Hendrycks and Dietterich(2018)]{hendrycks2018benchmarking}
Dan Hendrycks and Thomas~G Dietterich.
\newblock Benchmarking neural network robustness to common corruptions and
  surface variations.
\newblock \emph{arXiv preprint arXiv:1807.01697}, 2018.

\bibitem[LeCun(1998)]{lecun1998mnist}
Yann LeCun.
\newblock The mnist database of handwritten digits.
\newblock \emph{http://yann. lecun. com/exdb/mnist/}, 1998.

\bibitem[Deng(2012)]{deng2012mnist}
Li~Deng.
\newblock The mnist database of handwritten digit images for machine learning
  research [best of the web].
\newblock \emph{IEEE signal processing magazine}, 29\penalty0 (6):\penalty0
  141--142, 2012.

\bibitem[Bulatov(2011)]{bulatov2011notmnist}
Yaroslav Bulatov.
\newblock Notmnist dataset.
\newblock \emph{Google (Books/OCR), Tech. Rep.[Online]. Available:
  http://yaroslavvb. blogspot. it/2011/09/notmnist-dataset. html}, 2, 2011.

\bibitem[Gatys et~al.(2015)Gatys, Ecker, and Bethge]{gatys2015texture}
Leon Gatys, Alexander~S Ecker, and Matthias Bethge.
\newblock Texture synthesis using convolutional neural networks.
\newblock \emph{Advances in neural information processing systems}, 28, 2015.

\bibitem[Soriano et~al.(1996)Soriano, Spillmann, and Bach]{soriano1996abutting}
Manuel Soriano, Lothar Spillmann, and Michael Bach.
\newblock The abutting grating illusion.
\newblock \emph{Vision research}, 36\penalty0 (1):\penalty0 109--116, 1996.

\bibitem[Song and Baker(2006)]{song2006neural}
Yuning Song and Curtis~L Baker.
\newblock Neural mechanisms mediating responses to abutting gratings: luminance
  edges vs. illusory contours.
\newblock \emph{Visual neuroscience}, 23\penalty0 (2):\penalty0 181--199, 2006.

\bibitem[Marcel and Rodriguez(2010)]{marcel2010torchvision}
S{\'e}bastien Marcel and Yann Rodriguez.
\newblock Torchvision the machine-vision package of torch.
\newblock In \emph{Proceedings of the 18th ACM international conference on
  Multimedia}, pages 1485--1488, 2010.

\bibitem[Miller(1995)]{miller1995wordnet}
George~A Miller.
\newblock Wordnet: a lexical database for english.
\newblock \emph{Communications of the ACM}, 38\penalty0 (11):\penalty0 39--41,
  1995.

\bibitem[Kolesnikov et~al.(2020)Kolesnikov, Beyer, Zhai, Puigcerver, Yung,
  Gelly, and Houlsby]{kolesnikov2020big}
Alexander Kolesnikov, Lucas Beyer, Xiaohua Zhai, Joan Puigcerver, Jessica Yung,
  Sylvain Gelly, and Neil Houlsby.
\newblock Big transfer (bit): General visual representation learning.
\newblock In \emph{European conference on computer vision}, pages 491--507.
  Springer, 2020.

\bibitem[Ridnik et~al.(2021)Ridnik, Ben-Baruch, Noy, and
  Zelnik-Manor]{ridnik2021imagenet}
Tal Ridnik, Emanuel Ben-Baruch, Asaf Noy, and Lihi Zelnik-Manor.
\newblock Imagenet-21k pretraining for the masses.
\newblock \emph{arXiv preprint arXiv:2104.10972}, 2021.

\bibitem[Codreanu et~al.(2017)Codreanu, Podareanu, and
  Saletore]{codreanu2017scale}
Valeriu Codreanu, Damian Podareanu, and Vikram Saletore.
\newblock Scale out for large minibatch sgd: Residual network training on
  imagenet-1k with improved accuracy and reduced time to train.
\newblock \emph{arXiv preprint arXiv:1711.04291}, 2017.

\bibitem[Wong et~al.(2020)Wong, Rice, and Kolter]{wong2020fast}
Eric Wong, Leslie Rice, and J~Zico Kolter.
\newblock Fast is better than free: Revisiting adversarial training.
\newblock \emph{arXiv preprint arXiv:2001.03994}, 2020.

\bibitem[DeVries and Taylor(2017)]{devries2017improved}
Terrance DeVries and Graham~W Taylor.
\newblock Improved regularization of convolutional neural networks with cutout.
\newblock \emph{arXiv preprint arXiv:1708.04552}, 2017.

\bibitem[Zhang et~al.(2017)Zhang, Cisse, Dauphin, and
  Lopez-Paz]{zhang2017mixup}
Hongyi Zhang, Moustapha Cisse, Yann~N Dauphin, and David Lopez-Paz.
\newblock mixup: Beyond empirical risk minimization.
\newblock \emph{arXiv preprint arXiv:1710.09412}, 2017.

\bibitem[Verma et~al.(2019)Verma, Lamb, Beckham, Najafi, Mitliagkas, Lopez-Paz,
  and Bengio]{verma2019manifold}
Vikas Verma, Alex Lamb, Christopher Beckham, Amir Najafi, Ioannis Mitliagkas,
  David Lopez-Paz, and Yoshua Bengio.
\newblock Manifold mixup: Better representations by interpolating hidden
  states.
\newblock In \emph{International Conference on Machine Learning}, pages
  6438--6447. PMLR, 2019.

\bibitem[Yun et~al.(2019)Yun, Han, Oh, Chun, Choe, and Yoo]{yun2019cutmix}
Sangdoo Yun, Dongyoon Han, Seong~Joon Oh, Sanghyuk Chun, Junsuk Choe, and
  Youngjoon Yoo.
\newblock Cutmix: Regularization strategy to train strong classifiers with
  localizable features.
\newblock In \emph{Proceedings of the IEEE/CVF international conference on
  computer vision}, pages 6023--6032, 2019.

\bibitem[Hooker et~al.(2019)Hooker, Courville, Clark, Dauphin, and
  Frome]{hooker2019compressed}
Sara Hooker, Aaron Courville, Gregory Clark, Yann Dauphin, and Andrea Frome.
\newblock What do compressed deep neural networks forget?
\newblock \emph{arXiv preprint arXiv:1911.05248}, 2019.

\bibitem[Hendrycks et~al.(2021{\natexlab{b}})Hendrycks, Zhao, Basart,
  Steinhardt, and Song]{hendrycks2021natural}
Dan Hendrycks, Kevin Zhao, Steven Basart, Jacob Steinhardt, and Dawn Song.
\newblock Natural adversarial examples.
\newblock In \emph{Proceedings of the IEEE/CVF Conference on Computer Vision
  and Pattern Recognition}, pages 15262--15271, 2021{\natexlab{b}}.

\bibitem[Selvaraju et~al.(2017)Selvaraju, Cogswell, Das, Vedantam, Parikh, and
  Batra]{selvaraju2017grad}
Ramprasaath~R Selvaraju, Michael Cogswell, Abhishek Das, Ramakrishna Vedantam,
  Devi Parikh, and Dhruv Batra.
\newblock Grad-cam: Visual explanations from deep networks via gradient-based
  localization.
\newblock In \emph{Proceedings of the IEEE international conference on computer
  vision}, pages 618--626, 2017.

\bibitem[Sundararajan et~al.(2017)Sundararajan, Taly, and
  Yan]{sundararajan2017axiomatic}
Mukund Sundararajan, Ankur Taly, and Qiqi Yan.
\newblock Axiomatic attribution for deep networks.
\newblock In \emph{International conference on machine learning}, pages
  3319--3328. PMLR, 2017.

\bibitem[Bravo et~al.(1988)Bravo, Blake, and Morrison]{bravo1988cats}
Mary Bravo, Randolph Blake, and Sharon Morrison.
\newblock Cats see subjective contours.
\newblock \emph{Vision research}, 28\penalty0 (8):\penalty0 861--865, 1988.

\end{thebibliography}






\end{document}